\pdfoutput=1
\documentclass{article}
\PassOptionsToPackage{dvipsnames,table}{xcolor}
\usepackage{aibl_arxiv}

\usepackage{amsmath,amssymb}
\usepackage{graphicx}
\usepackage[dvipsnames,table]{xcolor}
\usepackage{colortbl}
\usepackage{booktabs}
\usepackage{multirow}
\usepackage{makecell}
\usepackage{pifont}
\usepackage{enumitem}
\usepackage{algorithm}
\usepackage{algorithmic}
\usepackage[capitalize,nameinlink,noabbrev]{cleveref}
\graphicspath{{figures/}}

\usepackage{etoolbox}
\apptocmd{\thebibliography}{\clubpenalty=0 \widowpenalty=0 \interlinepenalty=0}{}{}

\renewcommand\aiblresource[2]{{\sffamily\bfseries\small #1:} {\footnotesize\mbox{\url{#2}}}\par}

\makeatletter
\renewcommand\section{\@startsection{section}{1}{\z@}%
  {-2.0ex \@plus -1ex \@minus -.2ex}{1.0ex \@plus .2ex}%
  {\sffamily\bfseries\large\color{aiblteal}\nobreak}}
\renewcommand\subsection{\@startsection{subsection}{2}{\z@}%
  {-1.6ex \@plus -1ex \@minus -.2ex}{0.8ex \@plus .2ex}%
  {\sffamily\bfseries\normalsize\color{aiblteal}\nobreak}}
\makeatother

\usepackage{accsupp}
\newcommand\affillogo[3][]{%
  \if\relax\detokenize{#1}\relax
    \raisebox{\dimexpr 0.33em-0.5\height\relax}{\includegraphics[height=#3]{#2}}%
  \else
    \BeginAccSupp{method=escape,ActualText={#1}}\raisebox{\dimexpr 0.33em-0.5\height\relax}{\includegraphics[height=#3]{#2}}\EndAccSupp{}%
  \fi}

\newcommand{\cmark}{\textcolor{ForestGreen}{\ding{51}}}
\newcommand{\xmark}{\textcolor{BrickRed}{\ding{55}}}
\newcommand{\ub}[1]{\textcolor{gray}{#1}}
\DeclareMathOperator*{\argmax}{arg\,max}
\DeclareMathOperator*{\argmin}{arg\,min}

\crefname{figure}{Fig.}{Figs.}
\Crefname{figure}{Figure}{Figures}
\crefname{table}{Tab.}{Tabs.}
\Crefname{table}{Table}{Tables}
\crefname{section}{Sec.}{Secs.}
\Crefname{section}{Section}{Sections}
\crefname{equation}{Eq.}{Eqs.}
\Crefname{equation}{Equation}{Equations}
\crefname{algorithm}{Alg.}{Algs.}
\Crefname{algorithm}{Algorithm}{Algorithms}



\begin{document}
\clubpenalty=10000
\widowpenalty=10000
\displaywidowpenalty=10000

\aiblmaketitle
  {One Click per Cell Type Suffices: Training-free Group Interaction for Cell Instance Segmentation}
  {\fitbox{Sanghyun Jo$^{1,2}$, Seo Jin Lee$^{1}$, Seohyung Hong$^{1}$, Yoorim Gang$^{1}$, Hyeongsub Kim$^{1,3}$, Hyungseok Seo$^{1,\dagger}$, Kyungsu Kim$^{1,\dagger}$}}
  {$^{1}$\,\affillogo{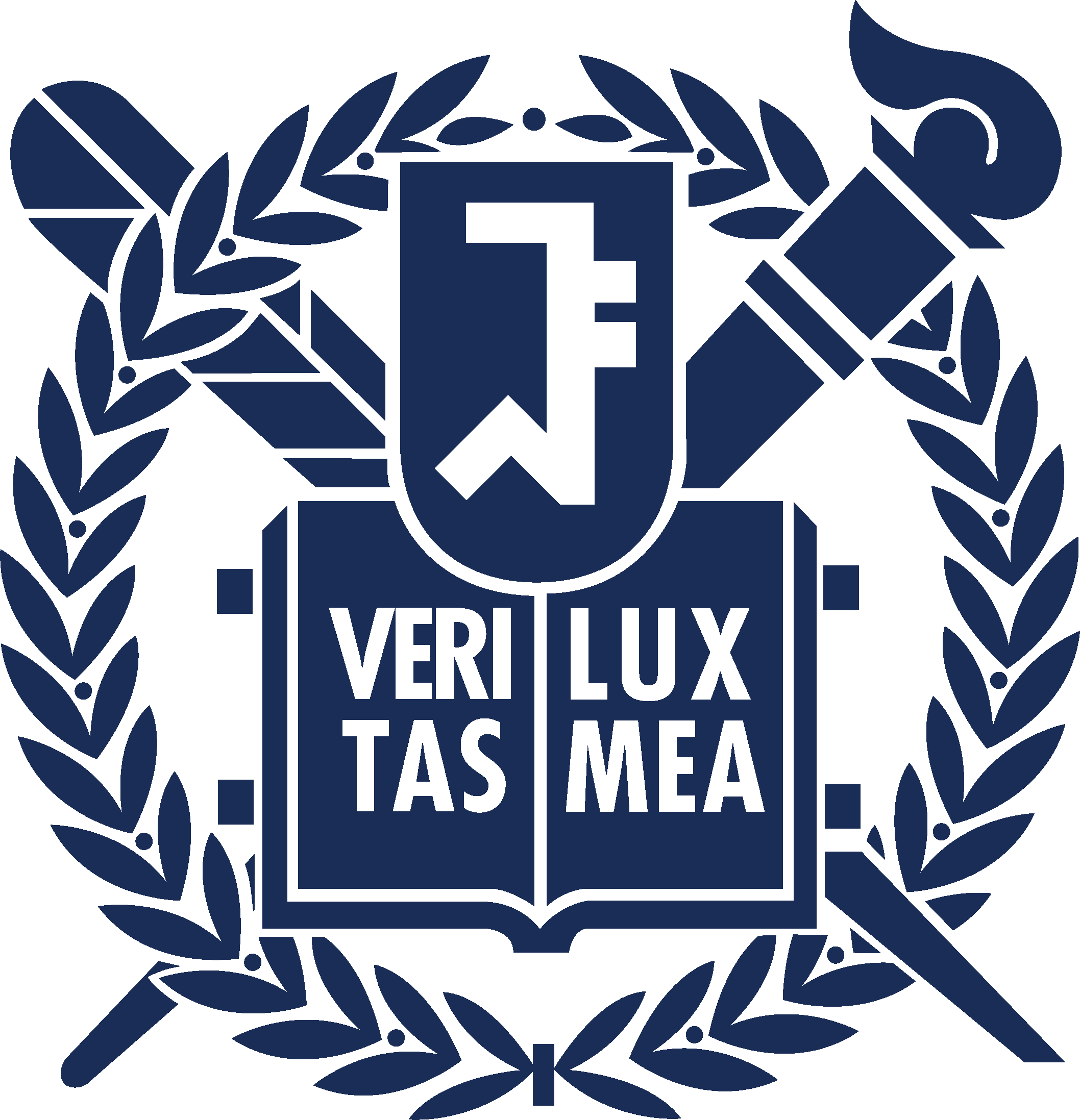}{15pt}\,Seoul National University \qquad $^{2}$\,\affillogo[OGQ]{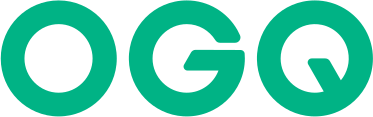}{8.6pt} \qquad $^{3}$\,\affillogo[LG CNS]{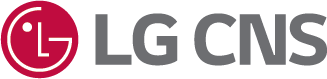}{10.4pt}\\[0.3em]
   {\small $^{\dagger}$Corresponding authors: \texttt{h.seo@snu.ac.kr}, \texttt{kyskim@snu.ac.kr}}}
  {Cell instance segmentation models trained on cell-specific datasets suffer severe performance drops on out-of-distribution cell types, while interactive foundation models overcome this through per-instance prompting at a cost that is prohibitively expensive for histopathology images containing hundreds to thousands of densely packed instances. We introduce \textbf{Group Prompting}, which shifts interactive segmentation from per-instance $O(N)$ to per-type $O(T)$, where a single click per cell type suffices to segment all instances of that type. Our key observation is that the frozen image encoder of the Segment Anything Model (SAM) already clusters same-type cells in its feature space before any prompt is given, and that this clustering holds across staining modalities without any training. Exploiting this property, we propose \textbf{Chain-of-Prompts (CoP)}, a training-free framework that recursively expands a single user click by (1) identifying reliable same-type locations through non-parametric gating of multi-scale encoder features, and (2) selecting the most spatially distant reliable point as the next prompt to maximize coverage. On eleven benchmarks, CoP generalizes to both unseen cell types and unseen imaging modalities without any adaptation: with one click per type it retains over 90\% of SAM3's per-instance performance on three cell-type-annotated datasets while surpassing fully-supervised methods, and with one click per image it retains over 95\% on eight datasets spanning both H\&E and non-H\&E imaging.}
  {\aiblresource{Project Page}{https://shjo-april.github.io/Chain-of-Prompts/}
   \aiblresource{Github}{https://github.com/shjo-april/Chain-of-Prompts}
   \aiblresource{Lab}{https://aibl.snu.ac.kr}}
\suppressfloats[t]

\section{Introduction}
\label{sec:intro}

\begin{figure*}[t]
  \centering
  \includegraphics[width=0.84\linewidth]{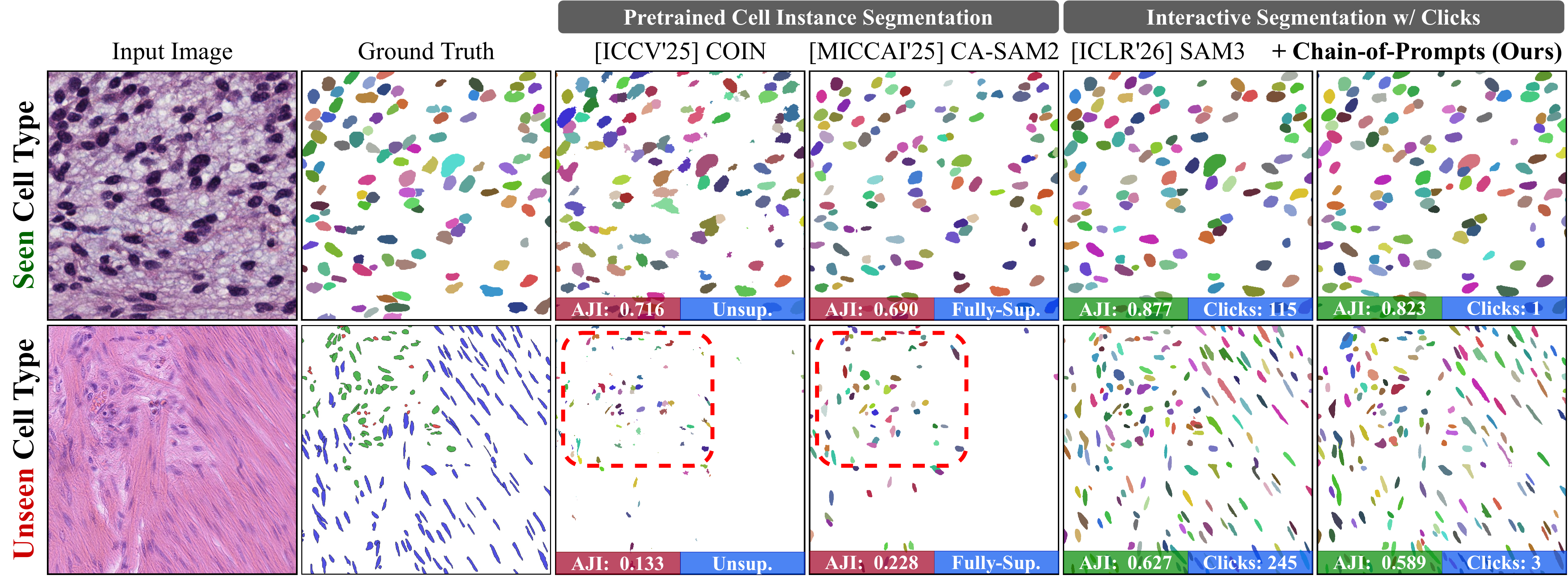}
  \caption{
    \textbf{One Click per Cell Type is All You Need.}
    {Pretrained models fail to identify unseen types and their performance is limited to a specific cell type (\textcolor{red}{dashed boxes}). While SAM3 \cite{SAM3} generalizes, it requires per-instance clicks (\emph{e.g.,} 245). Our CoP achieves 93.9\% of the upper bound performance \cite{SAM3} with only 3 clicks.}
  }
  \label{fig_teaser}
\end{figure*}

Cell instance segmentation is essential for quantitative analysis in computational pathology, yet existing cell-specific methods \cite{SSA,PSM} remain fundamentally constrained by their training data. Whether unsupervised \cite{COIN}, weakly-supervised \cite{DES-SAM}, or fully-supervised \cite{CA-SAM2}, these approaches learn cell representations tied to specific tissue types and cell morphologies encountered during training, leading to severe performance degradation on out-of-distribution (OOD) cell types (see Fig.~\ref{fig_teaser}). Recent interactive foundation models such as SAM3 \cite{SAM3} offer an alternative by accepting per-instance point prompts, enabling segmentation of arbitrary cell types without task-specific training. However, unlike natural images \cite{VOC,COCO} where object numbers are in the tens, histopathology images \cite{TNBC,CoNSeP} contain hundreds to thousands of densely packed cell instances, making per-instance prompting prohibitively expensive in practice. This contrast motivates a paradigm shift from per-instance prompting, which scales as $O(N)$ with the number of cells, to per-type group prompting at $O(T)$, where a single click per cell type suffices to segment all instances of that type.

{A common strategy to reduce per-instance cost is to generate pseudo prompts (\emph{e.g.,} points) automatically using external open-vocabulary or cell-specific detection models \cite{G-DINO,YOLOE,Rex-Omni}. However, these detectors are trained on specific cell and tissue types and therefore inherit the same OOD limitation (see Fig.~\ref{fig_teaser}). In this work, we bypass external detectors by leveraging a key intrinsic property of SAM \cite{SAM1,SAM3,uSAM}. Because SAM's architecture dictates that the image encoder must embed all instance information before receiving user prompts at the decoding stage, its frozen feature space inherently performs instance-aware encoding. Combined with the morphological traits that cells of one type share, such as size, shape and staining pattern, this yields cell-type-aware clustering without supervision, so similarity to one cell's feature already picks out the rest of its type.}

{While this property grounds the propagation of a single click to all instances of the same cell type, exploiting it presents two challenges.} First, SAM's multi-scale features dictate a strict trade-off between spatial precision and type selectivity: high-resolution features localize densely but activate background regions with similar texture, whereas low-resolution features accurately isolate cell types but blur adjacent instances. Second, one-shot propagation~\cite{PerSAM,Matcher} is threshold-sensitive, giving false positives or missed cells depending on where the cutoff is placed.

\begin{figure*}[t]
  \centering
  \includegraphics[width=0.78\linewidth]{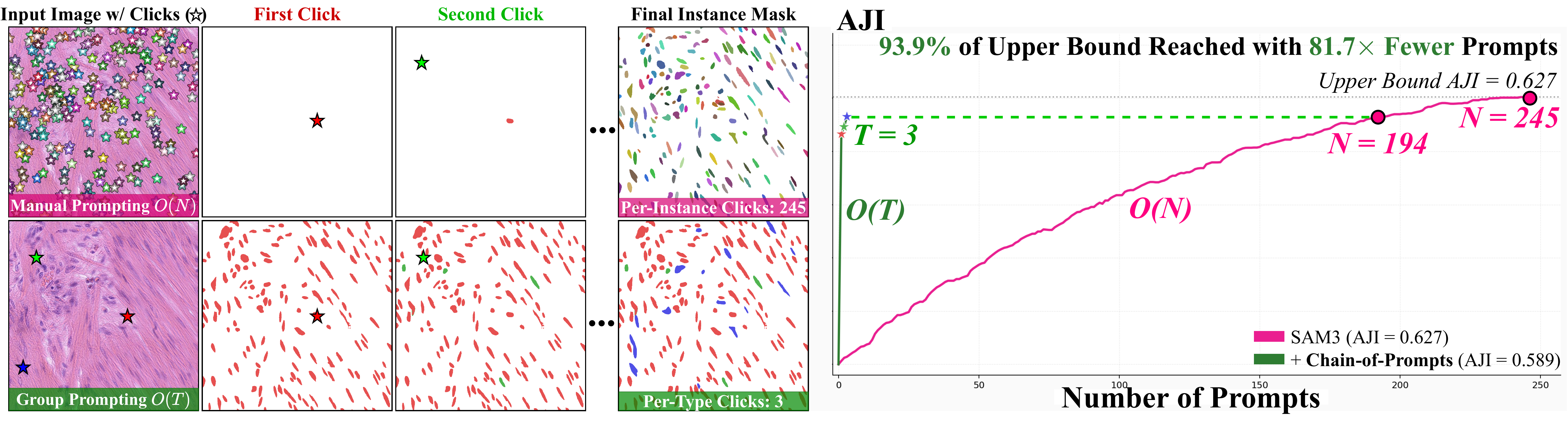}
  \caption{
    \textbf{From 245 Clicks to 3: Group Prompting.} Manual prompting requires one click per instance; our group prompting propagates each click to all same-type instances, reaching \textbf{93.9\%} of the upper bound with $\mathbf{81.7\times}$ fewer prompts.
  }
  \label{fig_highlight}
\end{figure*}

To address these challenges, we propose \textbf{Chain-of-Prompts (CoP)}, a training-free framework {that recursively leverages newly discovered cells as prompts for subsequent propagation.} CoP consists of two complementary components. First, Hierarchical Similarity Gating (HSG) combines SAM's multi-scale features to non-parametrically identify reliable cell points {recursively}, achieving precision above 96\% without any learnable parameters. Second, Farthest Prompt Recursion (FPR) {ensures comprehensive tissue coverage by selecting the next prompt furthest from all prior clicks, maximizing spatial diversity by uncovering cells in unexplored regions}. By iterating these two steps, our CoP expands from a single click to segment most of the same-type cells. On three benchmarks~\cite{CoNSeP,CoNIC,PanNuke},
{CoP uses only $O(T)$ per-type clicks and retains over 90\% of SAM3's \cite{SAM3} per-instance performance with 96.1\% reduction in annotation cost, while outperforming fully-supervised models \cite{CellViT,CA-SAM2,CellPose3}} (Fig.~\ref{fig_highlight}).

Our contributions are as follows:
\begin{itemize}[leftmargin=*,nosep]
    \item We introduce \textbf{Group Prompting}, shifting interactive segmentation from per-instance $O(N)$ to per-type $O(T)$, thereby reducing annotation cost from the number of cells to the number of cell types while staying robust to out-of-distribution types without cell-specific training.
    \item We propose \textbf{Chain-of-Prompts (CoP)}, a training-free framework that recursively expands prompt coverage while maintaining high precision ($\geq$96\%) at each iteration.
    \item On eleven benchmarks, CoP retains over 90\% of per-instance performance with SAM3 on three cell-type-annotated datasets (\cref{tab-typed}) and over 95\% on eight untyped datasets (\cref{tab-wotype,tab-nonhe}) spanning both H\&E and non-H\&E imaging, generalizing to unseen cell types and imaging modalities while outperforming fully-supervised methods~\cite{CellViT,CA-SAM2} trained on instance-level annotations.
\end{itemize}

\section{Related Work}
\label{sec:related}

\subsection{Interactive Segmentation Foundation Models}
The Segment Anything Model (SAM)~\cite{SAM1} recasts instance segmentation as a prompt-driven task that decodes one object from each user point or box. SAM2~\cite{SAM2} adds a memory module for image-to-video propagation, $\mu$SAM~\cite{uSAM} fine-tunes SAM on microscopy prompt-mask pairs to sharpen cell boundaries, and SAM3~\cite{SAM3} scales pretraining to medical data while extending the prompt vocabulary to text and visual exemplars. All inherit SAM's decomposed architecture, where a prompt-independent image encoder produces a feature map once and lightweight decoders turn each prompt into a mask. Point and box prompts therefore remain the most reliable pathway for cell-level quality, yet they fix the interaction cost at $O(N)$ since segmenting $N$ instances still requires $N$ prompts, which becomes prohibitive in dense histopathology where a single field can hold thousands of nuclei.

What per-instance usage overlooks is that prompt-mask training itself cultivates instance-level clustering in the encoder, so same-type cells are already grouped before any prompt is given, including for entirely unseen types. CoP exploits that latent structure to propagate one type-level click to every instance of that type without retraining.

\subsection{Cell-Specific Segmentation Models}
Existing methods aim to reduce annotation in cell instance segmentation (CIS), yet they trade scalability against robustness. Annotation-free approaches scale cheaply by deriving pseudo masks from self-supervised signals, where SSA~\cite{SSA} predicts patch magnification as self-supervision under the assumption that nucleus size and texture determine scale, PSM~\cite{PSM} aggregates shallow-layer gradients of a self-supervised network into prior self-activation maps and converts them into pseudo masks through semantic clustering, and COIN~\cite{COIN} suppresses pseudo-label noise via confidence-score-guided distillation. These priors hold for regular morphologies but break under shape variation and unseen cell types.

SPROUT~\cite{SPROUT} is training-free rather than annotation-free in the distillation sense: it derives foreground and background prototypes from H\&E stain priors and aligns features to them through a partial optimal transport schedule, then samples positive and negative point prompts for SAM. It is the closest training-free neighbor of CoP, sharing the idea of converting frozen features into SAM prompts, and differs on three axes. Its stain prior is specific to H\&E, so the non-H\&E benchmarks of \cref{tab-nonhe} are outside its scope, whereas CoP reads the same frozen encoder regardless of modality. It runs fully automatically, so the user cannot restrict the output to one cell population, whereas CoP spends one click per type precisely to obtain that control. And it requires a pathology foundation model for the prototype features in addition to the segmentation model it prompts, whereas CoP reads the propagation features out of the very encoder that will decode the masks.

Weakly-supervised SAM adaptations~\cite{DES-SAM,InstaSAM} still consume supervision that scales with the instance count: DES-SAM~\cite{DES-SAM}, for instance, attaches a segmentation head to a SAM backbone and distills it for box-supervised nuclear segmentation, using SAM only as a feature extractor behind a newly trained decoder. Learned prompt generators~\cite{PromptNucSeg,UN-SAM} remove the click only by training a prompter on nuclear instance annotations. Both families inherit the distribution of their training set, and neither propagates a prompt across instances. Fully-supervised models including CellViT~\cite{CellViT}, Cellpose3~\cite{CellPose3}, and CA-SAM2~\cite{CA-SAM2} attain strong in-domain quality but overfit to the annotated cell types and degrade sharply on out-of-distribution morphologies (\cref{tab-typed}), so reaching a new cell population costs another round of annotation and retraining. CoP stays training-free and label-free at test time, recovers all same-type instances from a single click, and still outperforms every one of these trained models on all three type-annotated benchmarks (\cref{tab-typed}), at one click per type rather than one per cell.

\subsection{Open-Vocabulary Segmentation Models}
A different route avoids per-instance clicks altogether, by conditioning the model on a category name or on one exemplar. Open-vocabulary models such as G-DINO~\cite{G-DINO}, YOLOE~\cite{YOLOE}, OpenWorldSAM~\cite{OpenWorldSAM}, and Rex-Omni~\cite{Rex-Omni}, the text and visual pathways of SAM3~\cite{SAM3}, and visual-prompt detectors and low-shot counters~\cite{T-Rex2,GeCo} return every region matching a text query or a reference exemplar from a single input. This yields near-$O(1)$ interaction, so the interface poses no obstacle; the difficulty lies in how the match is produced. These models depend on vision-language or exemplar alignment learned largely on natural images, where objects are few, spatially separated and morphologically distinct, and they optimize detection or counting rather than per-cell delineation. Such conditions rarely hold in dense, low-contrast cellular imagery, and their accuracy collapses under out-of-distribution pathology (\cref{tab-typed}), with text and visual prompts predicting reliably on only a subset of benchmarks and falling far short of point-prompted baselines. No existing route provides the match without domain-specific training while preserving per-cell granularity. CoP approaches the same efficiency through a different mechanism, issuing a single point prompt that directly queries the frozen image encoder whose features cluster same-type cells regardless of domain (\cref{fig:tsne}), then propagating it to all same-type instances via non-parametric similarity gating (HSG) and farthest-point recursion (FPR). This preserves the boundary precision of point prompting while approaching the annotation cost of open-vocabulary methods, shifting the paradigm from $O(N)$ to $O(T)$ without architectural modification or cell-specific retraining.

\subsection{Exemplar-Conditioned Segmentation}
PerSAM~\cite{PerSAM} and Matcher~\cite{Matcher} personalize a frozen model from one annotated reference without training. PerSAM~\cite{PerSAM} reads the reference feature out of SAM's own image encoder, takes the $\argmax$ of the resulting similarity map as a positive location prior and the $\argmin$ as a negative one, and refines the decoded mask in a cascade. Matcher~\cite{Matcher} instead matches patches with a separate DINOv2~\cite{DINOv2} encoder, samples point prompts by clustering the matched patches, and selects and merges the returned mask fragments. Both are formulated for cross-image one-shot transfer between a reference image and a different target image, neither addresses cells or dense instance layouts, and both emit semantic-granularity output: PerSAM~\cite{PerSAM} takes a single $\argmax$ location prior and Matcher~\cite{Matcher} merges its selected fragments, so a $T$-click budget caps the achievable instance count at $T$ no matter how many nuclei are present.

PerCS~\cite{PerCS} defines the reference-guided per-type cell segmentation task and solves it by integrating image features and reference embeddings through a cross-attention transformer over a self-supervised DINOv2~\cite{DINOv2} backbone, trained with contrastive supervision on a reorganized corpus of cell-type-annotated instance masks. Its instance separation is therefore learned from domain-specific ground-truth annotations and must be retrained for a new cell population, whereas CoP performs no gradient update and consumes no annotation at test time.

CoP shares with these works only the primitive of reading similarity out of frozen features. It differs in operating inside a single image from one click, in gating two feature scales against each other with a distribution-derived cutoff rather than a fixed ratio or a learned head, in recursing to the spatially farthest reliable point so that coverage is discovered instead of set by a cluster count, and in emitting one mask per discovered cell without any annotated instance.

\section{Method}

\begin{figure*}[t]
  \centering
  \includegraphics[width=0.78\linewidth]{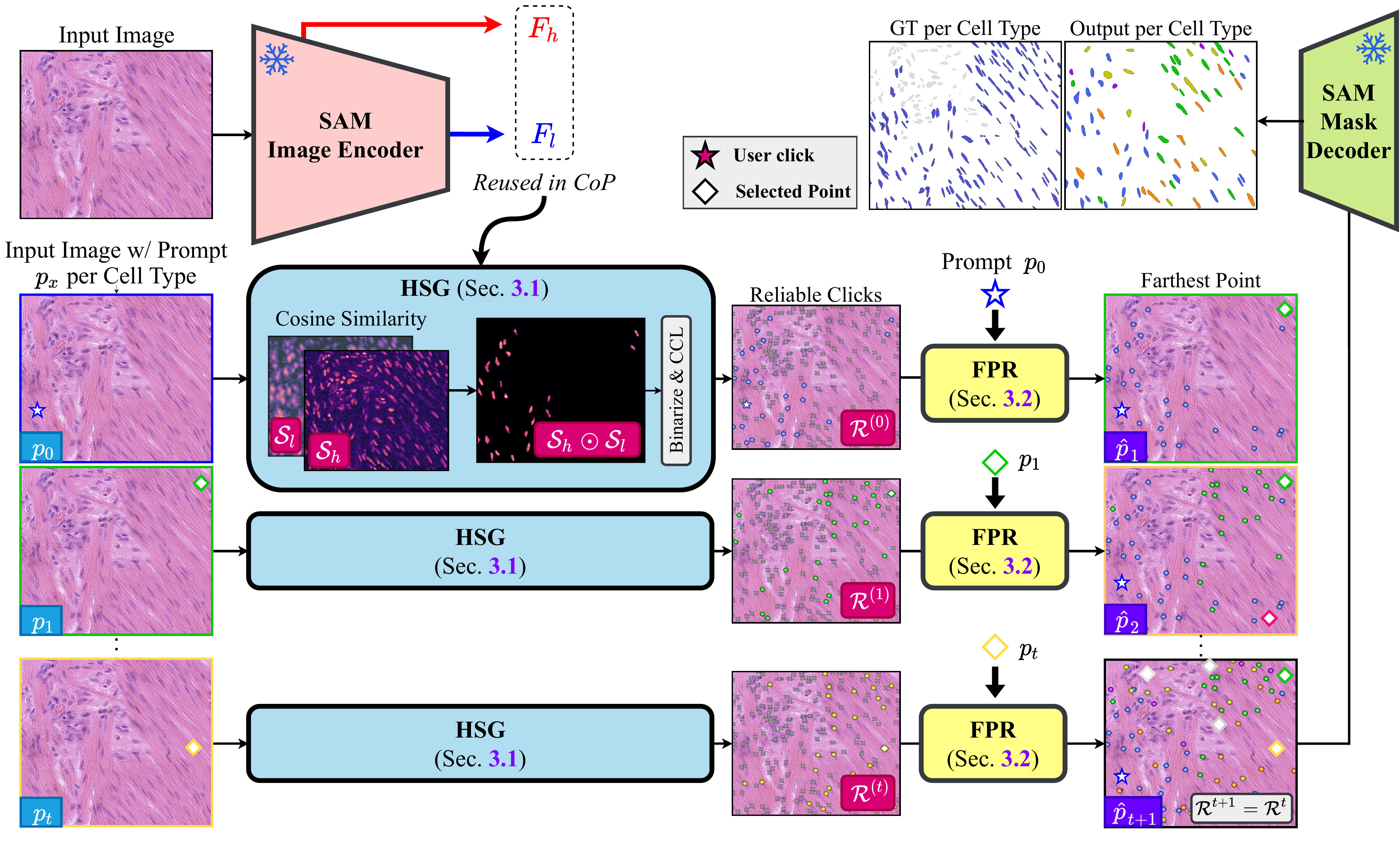}
  \caption{
    \textbf{Overview of Chain-of-Prompts (CoP).}
    A frozen SAM encoder extracts $F_h$ and $F_l$ once per image.
    For each user click $p_x$ (\ding{73}), HSG (\cref{sec:hsg}) produces initial reliable points $\mathcal{R}^{(0)}$ via hierarchical similarity and connected-component labeling (CCL).
    FPR (\cref{sec:fpr}) then expands $\mathcal{R}^{(0)}$ by iteratively prompting the farthest uncovered point (\ding{117}) until no new points are found. All propagated points per cell type are finally decoded into instance masks.
  }
  \label{fig_overview}
\end{figure*}

The proposed \textbf{Chain-of-Prompts (CoP)} is a training-free framework that discovers all same-type cells from a single user click and produces their instance masks. CoP operates exclusively on the frozen features of a pretrained SAM image encoder (\emph{e.g.,} SAM3~\cite{SAM3}), which extracts a high-resolution feature map $F_h \in \mathbb{R}^{D \times H/4 \times W/4}$ and a low-resolution feature map $F_l \in \mathbb{R}^{D \times H/16 \times W/16}$ from an input image $I$. As illustrated in Fig.~\ref{fig_overview}, CoP comprises two components. First, Hierarchical Similarity Gating (\cref{sec:hsg}) leverages the complementary strengths of $F_h$ and $F_l$ to identify a high-precision set of reliable points $\mathcal{R}^{(0)}$ from the initial prompt. Second, Farthest Prompt Recursion (\cref{sec:fpr}) then iteratively selects new prompts from this reliable set to expand spatial coverage until convergence ($\mathcal{R}^{(t+1)}{=}\mathcal{R}^{(t)}$). The resulting point set is decoded into instance masks via SAM's mask decoder, and \cref{alg:cop} summarizes the overall procedure.

\subsection{Hierarchical Similarity Gating} \label{sec:hsg}
A single feature scale cannot simultaneously achieve spatial precision and type selectivity. $F_h$ localizes individual cells even among tightly packed neighbors, but also activates tissue regions with similar texture beyond the target cell type. Conversely, $F_l$ selectively responds to the target type, but its coarse resolution causes neighboring instances to merge. HSG addresses this trade-off by combining both scales via element-wise gating to obtain a reliable point set $\mathcal{R}$ with high precision.

Given a point prompt $p$ per cell type, we interpolate $F_l$ to match the spatial resolution of $F_h$ and compute two cosine similarity maps: $S_h(x) = \cos(F_h(x),\, F_h(p))$ and $S_l(x) = \cos(F_l(x),\, F_l(p))$. The element-wise product $S_h \odot S_l$ suppresses false activations in $S_h$ that fall outside the target cell type according to $S_l$, while preserving spatially precise responses (Fig.~\ref{fig_overview}, HSG). We then binarize the gated map with a non-parametric threshold $\tau = \mu(S_h \odot S_l) + \sigma(S_h \odot S_l)$, inspired by COIN~\cite{COIN}, and apply connected-component labeling (CCL) to extract the similarity-weighted centroid from each connected region, as it provides a simple and deterministic way to convert dense activations into discrete point prompts without additional hyperparameters. The resulting centroids form the reliable set $\mathcal{R}^{(0)} = \{c_1, \ldots, c_K\}$, which covers cells near the prompt but misses distant instances.

\subsection{Farthest Prompt Recursion} \label{sec:fpr}
{While HSG identifies highly reliable cells in the local vicinity of the prompt, feature similarity naturally decays across distant, morphologically diverse tissue regions. Consequently, a single prompt yields lower precision for distant cells. FPR addresses this by automatically selecting the point in $\mathcal{R}^{(t)}$ that is furthest from all previously used prompts $\mathcal{Q}^{(t)} = \{p_0, \ldots, p_t\}$ at each iteration $t$:}

\begin{equation}
\label{eq:fpr}
  p_{t+1} = \argmax_{c \in \mathcal{R}^{(t)}}\;
  \min_{q \in \mathcal{Q}^{(t)}} \|c - q\|_2.
\end{equation}

{The selection rule of \cref{eq:fpr} is the classical farthest-point strategy~\cite{FPS}; what differs here is that its candidate set is not a fixed pool but is regenerated by HSG after every pick, so coverage and reliability advance together. By computing distance in image coordinates rather than feature space, we ensure each new prompt explores spatially uncovered tissue region without feature drift.
The selected prompt $p_{t+1}$ is then fed back into HSG (\cref{sec:hsg}) as a new prompt. Newly discovered points from the next round of HSG are merged into the reliable set: $\mathcal{R}^{(t+1)} = \mathcal{R}^{(t)} \cup \text{HSG}(p_{t+1}, F_h, F_l)$. This cycle repeats until no new points are discovered ($\mathcal{R}^{(t+1)} = \mathcal{R}^{(t)}$), indicating that every target cell that shares feature similarity to the initial click instance has been identified. Finally, each point $r \in \mathcal{R}$ is decoded into an instance mask via SAM's mask decoder (\cref{fig_overview}).}

\begin{algorithm}[t]
\footnotesize
\caption{Chain-of-Prompts (CoP)}
\label{alg:cop}
\begin{algorithmic}[1]
\REQUIRE image $I$; one click $p_x$ per cell type; frozen SAM encoder $\mathcal{E}$; decoder $\mathcal{D}$
\ENSURE instance masks $\mathcal{M}$
\STATE $(F_h, F_l) \gets \mathcal{E}(I)$ \COMMENT{$\triangleright$ encode once, reused for all clicks}
\STATE $\mathcal{R} \gets \varnothing$
\FORALL{user click $p_x$}
    \STATE $p_0 \gets p_x$;\ \ $\mathcal{R}^{(0)} \gets \mathrm{HSG}(p_0, F_h, F_l)$;\ \ $\mathcal{Q}^{(0)} \gets \{p_0\}$ \COMMENT{$\triangleright$ HSG (\cref{sec:hsg})}
    \FOR{$t = 0, 1, 2, \ldots$}
        \STATE $p_{t+1} \gets \argmax_{c \in \mathcal{R}^{(t)}} \min_{q \in \mathcal{Q}^{(t)}} \lVert c - q \rVert_2$ \COMMENT{$\triangleright$ FPR (\cref{sec:fpr})}
        \STATE $\mathcal{R}^{(t+1)} \gets \mathcal{R}^{(t)} \cup \mathrm{HSG}(p_{t+1}, F_h, F_l)$ \COMMENT{$\triangleright$ HSG (\cref{sec:hsg})}
        \STATE $\mathcal{Q}^{(t+1)} \gets \mathcal{Q}^{(t)} \cup \{p_{t+1}\}$
        \STATE \textbf{if} $\mathcal{R}^{(t+1)} = \mathcal{R}^{(t)}$ \textbf{then break}
    \ENDFOR
    \STATE $\mathcal{R} \gets \mathcal{R} \cup \mathcal{R}^{(t+1)}$
\ENDFOR
\STATE $\mathcal{M} \gets \{\mathcal{D}(r) : r \in \mathcal{R}\}$ \COMMENT{$\triangleright$ decode each point into an instance mask}
\STATE \textbf{return} $\mathcal{M}$
\end{algorithmic}
\end{algorithm}

\section{Experiments}
\label{sec:experiments}

\subsection{Experimental Setup}
\label{sec:setup}

\subsubsection{Datasets and Interaction Cost}
\label{sec:datasets}
We evaluate CoP on eleven cell instance segmentation benchmarks, all on their official test splits, grouped by two orthogonal factors, the availability of cell-type annotations and the staining modality: \emph{H\&E type-annotated} (CoNIC~\cite{CoNIC}, CoNSeP~\cite{CoNSeP}, PanNuke~\cite{PanNuke}), \emph{H\&E untyped} (MoNuSeg~\cite{MoNuSeg}, TNBC~\cite{TNBC}, CryoNuSeg~\cite{CryoNuSeg}, CPM-17~\cite{CPM-17}), and \emph{non-H\&E} (CellBinDB~\cite{CellBinDB}, Cellpose~\cite{Cellpose}, Kromp~\cite{Kromp}, LIVECell~\cite{LIVECell}), each reported in its own table (\cref{tab-typed,tab-wotype,tab-nonhe}).

\begin{table}[t]
\centering
\setlength{\tabcolsep}{5pt}
\caption{\textbf{Dataset statistics and interaction cost on the evaluation split.} \#Types is the number of annotated foreground classes; ``--'' denotes untyped datasets. $\mathcal{P}_N$ ($N$/img) and $\mathcal{P}_T$ ($T$/img) are the average per-image click counts of per-instance and per-type prompting, where $T$/img counts only the types present in each image. Click Reduction by CoP is $\mathcal{R}$ of Eq.~(\ref{eq:reduction}).}
\label{tab:datasets}
\fitbox{\begin{tabular}{l c r r r r}
\toprule
Dataset & \#Types & \#Images & \makecell{$\mathcal{P}_N$\\($N$/img)} & \makecell{$\mathcal{P}_T$\\($T$/img)} & \makecell{Click Reduction\\by CoP ($\mathcal{R}$)} \\
\midrule
\multicolumn{6}{l}{\textit{H\&E, cell-type annotated}} \\
CoNIC~\cite{CoNIC}     & 6 & 4{,}980 & 114.4 & 3.88 & 96.6\% \\
CoNSeP~\cite{CoNSeP}   & 7 & 14      & 626.9 & 3.86 & 99.4\% \\
PanNuke~\cite{PanNuke} & 5 & 2{,}722 & 24.5  & 2.05 & 91.6\% \\
\midrule
\multicolumn{6}{l}{\textit{H\&E, untyped}} \\
MoNuSeg~\cite{MoNuSeg}     & -- & 14      & 478.4 & 1.0 & 99.8\% \\
TNBC~\cite{TNBC}           & -- & 10      & 67.2  & 1.0 & 98.5\% \\
CryoNuSeg~\cite{CryoNuSeg} & -- & 10      & 242.4 & 1.0 & 99.6\% \\
CPM-17~\cite{CPM-17}       & -- & 32      & 118.9 & 1.0 & 99.2\% \\
\midrule
\multicolumn{6}{l}{\textit{non-H\&E, untyped}} \\
CellBinDB~\cite{CellBinDB} & -- & 303     & 80.5  & 1.0 & 98.8\% \\
Cellpose~\cite{Cellpose}   & -- & 68      & 105.9 & 1.0 & 99.1\% \\
Kromp~\cite{Kromp}         & -- & 37      & 137.5 & 1.0 & 99.3\% \\
LIVECell~\cite{LIVECell}   & -- & 1{,}512 & 297.6 & 1.0 & 99.7\% \\
\midrule
\textit{All (11 datasets)} & -- & 9{,}702 & 118.1 & 2.78 & \textbf{97.6\%} \\
\bottomrule
\end{tabular}}
\end{table}

\Cref{tab:datasets} summarizes these benchmarks, with all statistics measured on each dataset's evaluation split. The three typed datasets provide per-cell type labels and receive one click per type ($\mathcal{P}_T$); the remaining eight are untyped and receive a single click per image, as their instances are morphologically homogeneous and behave as one type. For each typed dataset we use all cell-type classes provided in its official release: CoNIC~\cite{CoNIC} with six (neutrophil, epithelial, lymphocyte, plasma, eosinophil, connective), CoNSeP~\cite{CoNSeP} with seven (other, inflammatory, healthy epithelial, malignant epithelial, fibroblast, muscle, endothelial), and PanNuke~\cite{PanNuke} with five (neoplastic, inflammatory, connective, dead, epithelial). The eight untyped datasets span H\&E histopathology (MoNuSeg~\cite{MoNuSeg}, TNBC~\cite{TNBC}, CryoNuSeg~\cite{CryoNuSeg}, CPM-17~\cite{CPM-17}) and non-H\&E microscopy: fluorescence (CellBinDB~\cite{CellBinDB}), phase-contrast (LIVECell~\cite{LIVECell}), fluorescence (Kromp~\cite{Kromp}), and varied modalities (Cellpose~\cite{Cellpose}). For CellBinDB~\cite{CellBinDB} we use the DAPI subset, and PanNuke is evaluated on its last fold (Fold~3) following its official split.

Group prompting shifts interaction from the per-instance $O(N)$ of conventional point (or box) prompting ($\mathcal{P}_N$) to the per-type $O(T)$ of CoP ($\mathcal{P}_T$). For image $k$ in a split of $K$ images, let $N_k$ be its number of instances and $T_k$ the number of cell types actually present in it; per-instance prompting places $N_k$ clicks while CoP places only $T_k$ and propagates each to all same-type instances. $T_k$ therefore measures how many visually distinct populations a field contains, not how many classes a user has to name: in practice one clicks each population once and stops when a new click returns nothing new, which is why the measured budget in \cref{tab:datasets} (3.88, 3.86, and 2.05 clicks per image) is well below the 6, 7, and 5 classes that CoNIC~\cite{CoNIC}, CoNSeP~\cite{CoNSeP}, and PanNuke~\cite{PanNuke} annotate. We define the click reduction $\mathcal{R}$ as the fraction of clicks CoP removes relative to per-instance prompting, summed over all $K$ images of the split before taking the ratio (a micro-average):
\begin{equation}
\mathcal{R}=1-\frac{\sum_{k=1}^{K} T_k}{\sum_{k=1}^{K} N_k},
\label{eq:reduction}
\end{equation}
where empty patches ($N_k{=}T_k{=}0$) contribute to neither sum. Measured on the evaluation splits, CoP removes $\mathcal{R}=96.1\%$ of the clicks over the three typed benchmarks, $99.6\%$ over the eight untyped ones, and $97.6\%$ over all eleven ($42.5\times$ fewer clicks). The typed-dataset reduction is the figure stated in \cref{sec:intro}, and the conclusion reports the overall one. The saving holds across both regimes, confirming that a single type-level prompt scales to arbitrarily many same-type instances at no additional interaction cost.

\subsubsection{Evaluation Metrics}
\label{sec:metrics}
Following standard practice~\cite{CoNIC,CA-SAM2,SSA,COIN} we report the Aggregated Jaccard Index (AJI), which scores instance-level agreement and penalizes false positives, missed instances, and split or merge errors, and the Dice coefficient, which scores pixel-level foreground overlap. Both are computed over every annotated instance regardless of its class, since cell instance segmentation asks for each cell to be delineated and not for cells to be classified; cell types enter only through the prompt budget, because populations that differ sharply in stain and morphology cannot be represented by one prompt.

For a single image, let $G_1,\dots,G_N$ be its $N$ ground-truth cell instances and $P_1,\dots,P_M$ its $M$ predicted instances, where each instance is the set of pixels belonging to one cell and $|\cdot|$ counts pixels. Instances are decoded so that each distinct color is one instance. The foreground regions are the unions $G=\bigcup_i G_i$ and $P=\bigcup_j P_j$, and the overlap of two regions is $\mathrm{IoU}(A,B)=|A\cap B|\,/\,|A\cup B|$.

The Dice coefficient compares the two foreground regions only, and is therefore instance-agnostic (\emph{i.e.,} binary segmentation):
\begin{equation}
\mathrm{Dice}=\frac{2\,|G\cap P|+\varepsilon}{|G|+|P|+\varepsilon},\qquad \varepsilon=1 .
\end{equation}
It rewards correct cell-versus-background delineation but does not penalize splitting one cell into several predictions or merging several cells into one, as long as the total foreground area agrees.

AJI instead scores matched instances and penalizes false positives. It first matches each ground-truth instance $G_i$ to the predicted instance of maximum overlap,
\begin{equation}
\sigma(i)=\operatorname*{arg\,max}_{\,j\,:\,P_j\cap G_i\neq\varnothing}\ \mathrm{IoU}(G_i,P_j),
\end{equation}
a greedy assignment over the predictions that touch $G_i$. Writing $\mathcal{A}=\{\sigma(i)\}_{i=1}^{N}$ for the set of matched predictions, AJI then aggregates intersections and unions over the whole image:
\begin{equation}
\mathrm{AJI}=\frac{\displaystyle\sum_{i=1}^{N}\bigl|G_i\cap P_{\sigma(i)}\bigr|}
{\displaystyle\sum_{i=1}^{N}\bigl|G_i\cup P_{\sigma(i)}\bigr|+\sum_{j\,:\,P_j\notin\mathcal{A}}\bigl|P_j\bigr|} .
\end{equation}
Here a ground-truth instance with no overlapping prediction adds only its own area $|G_i|$ to the denominator as a miss, and every unmatched prediction $P_j$ adds its full area $|P_j|$ as a false positive, so a single ratio penalizes boundary error, under-segmentation, over-segmentation, misses, and spurious detections at once. Both metrics are computed per image and averaged over the images of a split, so each image weighs equally regardless of its cell count. Neither metric reads the type label, so \cref{tab-typed,tab-wotype,tab-nonhe} score instance segmentation alone; \cref{sec:leakage} revisits the same predictions with the cell type taken into account. We report both because Dice measures region delineation while AJI measures instance separation: a high-Dice, low-AJI gap signals good foreground coverage but poor separation of touching cells, which is exactly the hard case for single-click prompting, where one prompt has to reach many adjacent cells.

\textbf{Detection metrics.} AJI and Dice are both area-based, so a cell contributes in proportion to its pixel count and one ratio absorbs misses, false positives and boundary error together. To separate those effects we also report detection scores, which count instances rather than pixels (\cref{tab:detection}). A prediction $P_j$ and a ground truth $G_i$ form a match when $\mathrm{IoU}(G_i,P_j)>0.5$; since that threshold exceeds one half, no instance can match two partners, so the assignment is one-to-one and needs no tie-breaking rule. Writing $\mathrm{TP}$ for the matched pairs, $\mathrm{FP}$ for the unmatched predictions and $\mathrm{FN}$ for the unmatched ground truths,
\begin{equation}
\mathrm{Prec.}=\frac{\mathrm{TP}}{\mathrm{TP}+\mathrm{FP}},\qquad \mathrm{Rec.}=\frac{\mathrm{TP}}{\mathrm{TP}+\mathrm{FN}},\qquad F_1=\frac{2\,\mathrm{Prec.}\cdot\mathrm{Rec.}}{\mathrm{Prec.}+\mathrm{Rec.}} .
\end{equation}
The cell-type-aware variant of \cref{tab:detection} adds one condition to the match, that the cell type of $P_j$, read from the click that produced it, equals the cell type of $G_i$, and averages over the cell types present in an image before averaging over images, so a rare cell type weighs as much as a frequent one.

The two families answer different questions about the same masks. AJI states how much annotated area CoP recovers with the instances kept apart, but it cannot say whether a shortfall comes from cells the propagation never reached or from cells it reached under the wrong cell type, because neither contributes to its numerator. Precision and Recall split exactly that: Precision drops only when CoP emits a mask that no ground truth of the queried cell type accepts, and Recall drops only when annotated cells are left unsegmented. The two families also weigh errors differently, since a missed small nucleus costs a full count in Recall but only its own few pixels in AJI, which makes the detection scores the stricter test on dense images of small cells.

\subsubsection{Baselines and Protocol}
\label{sec:protocol}
Every baseline runs with its official code and publicly released weights under its intended prompting protocol, ensuring a fair comparison. Open-vocabulary methods~\cite{YOLOE,OpenWorldSAM,Rex-Omni} take ``cell'' as the text prompt ($\mathcal{T}$), and those that also support visual prompting, YOLOE~\cite{YOLOE} and Rex-Omni~\cite{Rex-Omni}, additionally receive a cropped cell patch as the reference image ($\mathcal{V}$). Interactive baselines~\cite{SAM1,SAM2,SAM3,uSAM} receive $N$ foreground clicks ($\mathcal{P}_N$), each placed at the centroid of a ground-truth instance mask. Fully-supervised methods~\cite{CellViT,CA-SAM2,CellPose3} ($\mathcal{M}$) use their released models trained on their respective datasets. CoP introduces no trainable parameters. On type-annotated data a user provides a single click per cell type ($\mathcal{P}_T$), and on untyped data a single click per image is sufficient; from this initial prompt, HSG and FPR automatically propagate coverage to all remaining same-type instances. The protocol does not assume that the user knows which cell types an image contains: the first click starts the propagation, the user then clicks any cell that is still unsegmented, and the loop ends when none are left, so $T$ is discovered from the image rather than read off a dataset label, and rare or previously unseen cell types are absorbed by the same loop. Because a real user may click anywhere within a target cell rather than at a fixed or optimal location, we report every CoP result, in both the typed and untyped settings, as the mean over at least 10 independent runs, each starting from a randomly sampled initial click. All reported numbers therefore reflect expected performance under practical, non-deterministic prompting rather than a single favorable initialization. The same protocol applies to every CoP row in the paper, and the spread over these runs stays small everywhere (\cref{sec:robustness}), so no reported number depends on a favorable click. Duplicate masks are handled in two stages: CCL merges the prompts that fall inside one cell (\cref{sec:hsg}), and standard NMS at IoU 0.5 is applied as the final post-processing. The second stage is close to a no-op, since disabling it changes \cref{tab-typed} AJI by less than $0.002$ on average, confirming that almost no overlapping masks survive CCL.

\textbf{Open-vocabulary baseline protocol.} Open-vocabulary baselines differ in the granularity of their output. Mask-native models, such as YOLOE~\cite{YOLOE}, OpenWorldSAM~\cite{OpenWorldSAM}, and the text and visual pathways of SAM3~\cite{SAM3}, return instance masks directly and are evaluated as is. Detectors that output only bounding boxes, such as Rex-Omni~\cite{Rex-Omni}, predict no masks; for these we obtain an instance mask by prompting interactive segmentation models, such as SAM3~\cite{SAM3}, with each predicted box. Every open-vocabulary result in \cref{tab-typed,tab-wotype,tab-nonhe} is therefore scored at the instance-mask level with the same AJI and Dice as the other baselines, so the comparison reflects segmentation quality, not output format.

\textbf{Runtime and memory.} All experiments run on a single NVIDIA RTX PRO 6000. We report runtime on SAM3~\cite{SAM3}, the heaviest interactive baseline~\cite{SAM1,SAM2,SAM3,uSAM} and the one with the best average performance across eleven benchmarks, so these figures upper-bound CoP's cost over the interactive encoders. On a $1000{\times}1000$ input, image encoding is a one-time cost of about 0.6\,s. On CoNSeP~\cite{CoNSeP}, HSG and FPR propagation take 1.08\,s per click, converging after roughly 15 FPR iterations, while mask decoding adds 4.74\,ms per discovered prompt. At 3.86 clicks per image, encoding plus decoding takes 3.29\,s and propagation adds 4.17\,s. The $O(T)$ of group prompting counts user interaction; the prompts CoP decodes internally still scale with the cells it finds, as detailed in \cref{sec:budget}. Because CoP runs entirely in frozen feature space without backpropagation and reuses the precomputed image features, its peak VRAM footprint remains nearly identical to that of the underlying interactive segmentation model, adding negligible memory overhead beyond the frozen image encoder.

\subsection{Comparison with State-of-the-art Approaches}
\label{sec:comparison}

\begin{table*}[t]
    \centering
    \caption{\textbf{H\&E benchmarks with cell-type annotations.}
    Prompt types: $\mathcal{T}$ text (``cell''), $\mathcal{V}$ visual (cell patch),
    \xmark~no prompt,
    $\mathcal{M}$ pixel-level supervision, $\mathcal{P}_N$ one point per instance,
    $\mathcal{P}_T$ one point per cell type.
    \ub{Gray} rows are per-instance upper bounds ($\mathcal{P}_N$). Each \colorbox{gray!15}{shaded} row is CoP applied to the base model directly above it. \textbf{Bold} and \underline{underline} mark the best and second-best in each column. CellViT~\cite{CellViT} and CA-SAM2~\cite{CA-SAM2} are in-domain on H\&E, Cellpose3~\cite{CellPose3} cross-domain.}
    \label{tab-typed}
    \setlength{\tabcolsep}{4pt}
    \renewcommand{\arraystretch}{1.05}
    \fitbox{\begin{tabular}{lc|cc|cc|cc}
        \toprule
        \multirow{2}{*}{\textbf{Method}} & \multirow{2}{*}{\textbf{Prompt}}
            & \multicolumn{2}{c|}{\textbf{CoNIC}}
            & \multicolumn{2}{c|}{\textbf{CoNSeP}}
            & \multicolumn{2}{c}{\textbf{PanNuke}} \\
        & & AJI $\uparrow$ & Dice $\uparrow$ & AJI $\uparrow$ & Dice $\uparrow$ & AJI $\uparrow$ & Dice $\uparrow$ \\
        \midrule
        YOLOE \cite{YOLOE} \venue{ICCV'25}               & $\mathcal{T}$ & 0.000 & 0.057 & 0.000 & 0.000 & 0.000 & 0.084 \\
        OpenWorldSAM \cite{OpenWorldSAM} \venue{NeurIPS'25} & $\mathcal{T}$ & 0.001 & 0.071 & 0.000 & 0.027 & 0.001 & 0.082 \\
        SAM3 \cite{SAM3} \venue{ICLR'26}                 & $\mathcal{T}$ & 0.450 & 0.696 & 0.000 & 0.000 & 0.558 & 0.775 \\
        Rex-Omni \cite{Rex-Omni} \venue{CVPR'26}         & $\mathcal{T}$ & 0.075 & 0.211 & 0.153 & 0.316 & 0.398 & 0.622 \\
        \midrule
        YOLOE \cite{YOLOE} \venue{ICCV'25}               & $\mathcal{V}$ & 0.028 & 0.110 & 0.000 & 0.000 & 0.141 & 0.323 \\
        SAM3 \cite{SAM3} \venue{ICLR'26}                 & $\mathcal{V}$ & 0.390 & 0.620 & 0.000 & 0.000 & 0.533 & 0.768 \\
        Rex-Omni \cite{Rex-Omni} \venue{CVPR'26}         & $\mathcal{V}$ & 0.075 & 0.207 & 0.126 & 0.348 & 0.408 & 0.654 \\
        \midrule
        SSA \cite{SSA} \venue{MICCAI'20}            & \xmark & 0.072 & 0.162 & 0.082 & 0.198 & 0.395 & 0.671 \\
        COIN \cite{COIN} \venue{ICCV'25}            & \xmark & 0.148 & 0.354 & 0.188 & 0.372 & 0.478 & 0.743 \\
        SPROUT \cite{SPROUT} \venue{ICML'26}            & \xmark & 0.271 & 0.537 & 0.264 & 0.514 & 0.507 & 0.678 \\
        \midrule
        CellViT \cite{CellViT} \venue{MedIA'24}            & $\mathcal{M}$ & 0.371 & 0.670 & 0.495 & 0.802 & 0.660 & 0.864 \\
        CA-SAM2 \cite{CA-SAM2} \venue{MICCAI'25}           & $\mathcal{M}$ & 0.269 & 0.561 & 0.382 & 0.700 & 0.542 & 0.768 \\
        Cellpose3 \cite{CellPose3} \venue{Nature Methods'25} & $\mathcal{M}$ & 0.173 & 0.333 & 0.158 & 0.354 & 0.243 & 0.404 \\
        \midrule
        \ub{$\mu$SAM \cite{uSAM} \venue{Nature Methods'25}} & \ub{$\mathcal{P}_N$} & \ub{0.759} & \ub{0.912} & \ub{0.785} & \ub{0.900} & \ub{0.789} & \ub{0.950} \\
        \rowcolor{gray!15}
        \textbf{$\mu$SAM + CoP (Ours)} & $\mathcal{P}_T$ & \underline{0.652} & \underline{0.791} & \textbf{0.712} & \underline{0.826} & \underline{0.734} & \underline{0.918} \\
        \ub{SAM3 \cite{SAM3} \venue{ICLR'26}} & \ub{$\mathcal{P}_N$} & \ub{0.801} & \ub{0.938} & \ub{0.716} & \ub{0.867} & \ub{0.830} & \ub{0.975} \\
        \rowcolor{gray!15}
        \textbf{SAM3 + CoP (Ours)} & $\mathcal{P}_T$ & \textbf{0.731} & \textbf{0.898} & \underline{0.653} & \textbf{0.831} & \textbf{0.782} & \textbf{0.932} \\
        \bottomrule
    \end{tabular}}
\end{table*}

\begin{table*}[t]
    \centering
    \caption{\textbf{H\&E benchmarks without cell-type annotations.} Since no cell-type labels are available, CoP begins from a single click per image and propagates it across all instances. Notation and baseline domains as in \cref{tab-typed}.}
    \label{tab-wotype}
    \setlength{\tabcolsep}{3pt}
    \renewcommand{\arraystretch}{1.05}
    \fitbox{\begin{tabular}{l|cc|cc|cc|cc}
        \toprule
        \multirow{2}{*}{\textbf{Method}}
            & \multicolumn{2}{c|}{\textbf{MoNuSeg}}
            & \multicolumn{2}{c|}{\textbf{TNBC}}
            & \multicolumn{2}{c|}{\textbf{CryoNuSeg}}
            & \multicolumn{2}{c}{\textbf{CPM-17}} \\
        & AJI & Dice & AJI & Dice & AJI & Dice & AJI & Dice \\
        \midrule
        CellViT \cite{CellViT} \venue{MedIA'24}            & 0.676 & 0.832 & 0.669 & 0.817 & 0.492 & 0.800 & 0.703 & 0.853 \\
        COIN \cite{COIN} \venue{ICCV'25}                   & 0.580 & 0.794 & 0.568 & 0.774 & 0.212 & 0.741 & 0.610 & 0.821 \\
        CA-SAM2 \cite{CA-SAM2} \venue{MICCAI'25}           & 0.645 & 0.807 & 0.644 & 0.799 & 0.468 & 0.781 & 0.657 & 0.819 \\
        Cellpose3 \cite{CellPose3} \venue{Nature Methods'25} & 0.316 & 0.499 & 0.224 & 0.365 & 0.140 & 0.303 & 0.326 & 0.479 \\
        SPROUT \cite{SPROUT} \venue{ICML'26}                   & 0.379 & 0.568 & 0.575 & 0.748 & 0.409 & 0.752 & 0.537 & 0.720 \\
        \midrule
        \ub{$\mu$SAM \cite{uSAM} \venue{Nature Methods'25}}
            & \ub{0.837} & \ub{0.913} & \ub{0.866} & \ub{0.928} & \ub{0.795} & \ub{0.899} & \ub{0.850} & \ub{0.921} \\
        \rowcolor{gray!15}
        \textbf{$\mu$SAM + CoP (Ours)}
            & \textbf{0.834} & \textbf{0.908} & \textbf{0.862} & \textbf{0.918} & \textbf{0.790} & \textbf{0.898} & \underline{0.840} & \underline{0.916} \\
        \ub{SAM3 \cite{SAM3} \venue{ICLR'26}}
            & \ub{0.802} & \ub{0.894} & \ub{0.860} & \ub{0.854} & \ub{0.786} & \ub{0.897} & \ub{0.851} & \ub{0.922} \\
        \rowcolor{gray!15}
        \textbf{SAM3 + CoP (Ours)}
            & \underline{0.795} & \underline{0.893} & \underline{0.858} & \underline{0.850} & \underline{0.773} & \underline{0.884} & \textbf{0.846} & \textbf{0.919} \\
        \bottomrule
    \end{tabular}}
\end{table*}

\begin{table*}[!t]
    \centering
    \caption{\textbf{Non-H\&E benchmarks.}
    These datasets use non-H\&E microscopy and also lack type labels, so CoP runs from one click per image as in \cref{tab-wotype}.
    Notation as in \cref{tab-typed}, with the domains reversed: Cellpose3~\cite{CellPose3} is in-domain here, CellViT~\cite{CellViT} and CA-SAM2~\cite{CA-SAM2} cross-domain.}
    \label{tab-nonhe}
    \setlength{\tabcolsep}{3pt}
    \renewcommand{\arraystretch}{1.05}
    \fitbox{\begin{tabular}{l|cc|cc|cc|cc}
        \toprule
        \multirow{2}{*}{\textbf{Method}}
            & \multicolumn{2}{c|}{\textbf{CellBinDB}}
            & \multicolumn{2}{c|}{\textbf{Cellpose}}
            & \multicolumn{2}{c|}{\textbf{Kromp}}
            & \multicolumn{2}{c}{\textbf{LIVECell}} \\
        & AJI & Dice & AJI & Dice & AJI & Dice & AJI & Dice \\
        \midrule
        CellViT \cite{CellViT} \venue{MedIA'24}            & 0.002 & 0.003 & 0.012 & 0.021 & 0.005 & 0.011 & 0.041 & 0.069 \\
        COIN \cite{COIN} \venue{ICCV'25}                   & 0.008 & 0.187 & 0.028 & 0.480 & 0.013 & 0.261 & 0.055 & 0.122 \\
        CA-SAM2 \cite{CA-SAM2} \venue{MICCAI'25}           & 0.384 & 0.601 & 0.148 & 0.245 & 0.123 & 0.187 & 0.179 & 0.326 \\
        Cellpose3 \cite{CellPose3} \venue{Nature Methods'25} & 0.619 & 0.801 & \underline{0.657} & \underline{0.862} & 0.769 & 0.900 & \underline{0.628} & \textbf{0.893} \\
        \midrule
        \ub{$\mu$SAM \cite{uSAM} \venue{Nature Methods'25}}
            & \ub{0.770} & \ub{0.871} & \ub{0.653} & \ub{0.827} & \ub{0.855} & \ub{0.917} & \ub{0.570} & \ub{0.800} \\
        \rowcolor{gray!15}
        \textbf{$\mu$SAM + CoP (Ours)}
            & \underline{0.758} & \underline{0.857} & 0.628 & 0.806 & \underline{0.837} & \underline{0.904} & 0.546 & 0.769 \\
        \ub{SAM3 \cite{SAM3} \venue{ICLR'26}}
            & \ub{0.831} & \ub{0.909} & \ub{0.819} & \ub{0.917} & \ub{0.888} & \ub{0.939} & \ub{0.704} & \ub{0.876} \\
        \rowcolor{gray!15}
        \textbf{SAM3 + CoP (Ours)}
            & \textbf{0.817} & \textbf{0.899} & \textbf{0.800} & \textbf{0.900} & \textbf{0.880} & \textbf{0.926} & \textbf{0.671} & \underline{0.842} \\
        \bottomrule
    \end{tabular}}
\end{table*}

\Cref{tab-typed} demonstrates why the two shortcuts that avoid per-instance interaction are unreliable in histopathology. Open-vocabulary text and visual prompts \cite{YOLOE,OpenWorldSAM,SAM3,Rex-Omni} fire only where their vision-language alignment happens to transfer: SAM3 \cite{SAM3} reaches 0.450 and 0.558 AJI on CoNIC \cite{CoNIC} and PanNuke \cite{PanNuke}, yet collapses to 0.000 on CoNSeP \cite{CoNSeP}, and the remaining detectors stay near zero on most benchmarks. Because this alignment ties text and visual queries to natural-image appearance, it transfers only when a dataset happens to resemble that distribution and misfires once the tissue type or stain differs, which is why the same model succeeds on one benchmark and predicts almost nothing on another. Annotation-free priors \cite{SSA,COIN} fail in the opposite way, segmenting a single foreground without resolving type structure, so they trail the point-prompted models throughout. Fully-supervised methods \cite{CellViT,CA-SAM2,CellPose3} are strong only inside their training domain: even the best of them, CellViT \cite{CellViT}, reaches just 0.371 AJI on CoNIC \cite{CoNIC}, roughly half of what CoP recovers. Against all three groups, CoP turns one click per type into instance masks that surpass every open-vocabulary and fully-supervised baseline on all three benchmarks, and with SAM3 \cite{SAM3} it retains over 90\% of its own per-instance upper bound on each (\emph{e.g.,} 0.731 vs.\ 0.801 AJI on CoNIC \cite{CoNIC}) at roughly one click per type instead of hundreds. Crucially, this lead holds on PanNuke \cite{PanNuke}, where CellViT \cite{CellViT} is trained in-distribution, showing that querying the frozen encoder directly (\cref{fig:tsne}) is more robust than either category-level conditioning or task-specific mask supervision.

When cell types are unlabeled (\cref{tab-wotype}), a single click per image is sufficient: FPR (\cref{sec:fpr}) propagates it across the whole image, so CoP retains roughly 98--99\% of the per-instance upper bound for both $\mu$SAM \cite{uSAM} and SAM3 \cite{SAM3} while still outperforming every fully-supervised baseline \cite{CellViT,CA-SAM2,CellPose3}. A change of imaging modality provides a more demanding test (\cref{tab-nonhe}): it removes the H\&E appearance that supervised methods were trained on, while the frozen encoder's general-purpose features remain available to CoP. In this setting, the H\&E-trained supervised models \cite{CellViT,CA-SAM2} fail almost completely; for example, CellViT \cite{CellViT} drops from 0.676 AJI on H\&E MoNuSeg \cite{MoNuSeg} to 0.002 on non-H\&E CellBinDB \cite{CellBinDB}, since its learned appearance prior does not carry over to an unfamiliar stain. CoP performs no adaptation and relies only on the frozen image encoder's features, and it preserves over 95\% of the per-instance upper bound on all four non-H\&E benchmarks \cite{CellBinDB,Cellpose,Kromp,LIVECell}. The one competitive trained model is Cellpose3 \cite{CellPose3}, whose training data covers these modalities: whereas it is the weakest supervised method on the H\&E benchmarks (\cref{tab-typed,tab-wotype}), it becomes the strongest trained baseline here and surpasses CoP only on LIVECell \cite{LIVECell} Dice (Cellpose3: 0.893 vs.\ CoP: 0.842); on every other column CoP outperforms every fully-trained baseline. Together, these two settings identify the image encoder's instance-level clustering, rather than any cell- or stain-specific training, as the property CoP depends on, and show that it transfers to both unseen cell types and unseen imaging modalities.

\begin{figure}[!t]
  \centering
  \includegraphics[width=\linewidth,height=0.53\textheight,keepaspectratio]{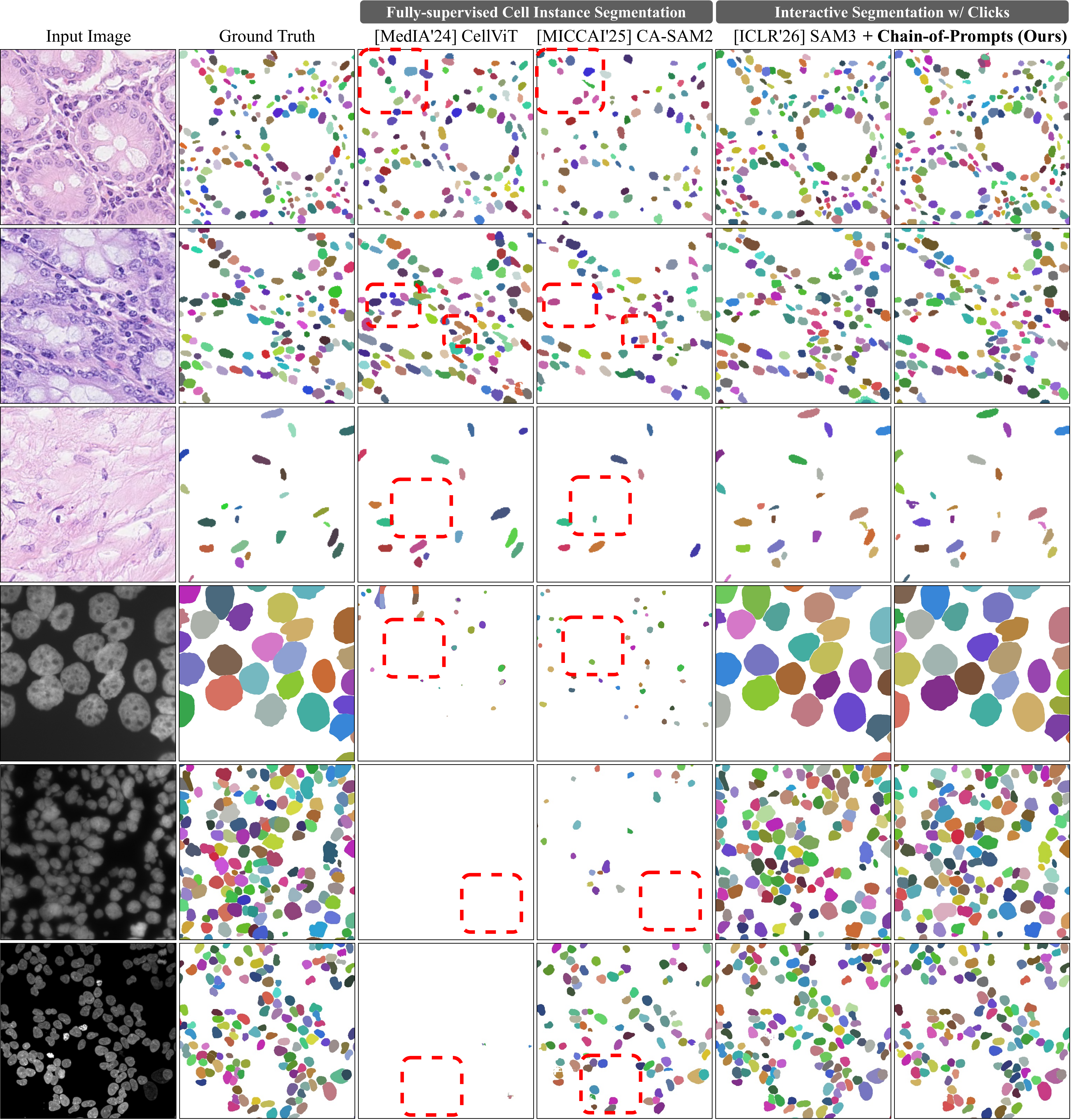}
  \caption{\textbf{Qualitative comparison across H\&E and non-H\&E benchmarks.} Each row is a dataset, spanning H\&E (\emph{e.g.,} CoNIC~\cite{CoNIC}) and non-H\&E (\emph{e.g.,} Kromp~\cite{Kromp}). Columns: input, ground truth, fully-supervised \cite{CellViT,CA-SAM2}, per-instance SAM3~\cite{SAM3} ($\mathcal{P}_N$), and SAM3 + CoP ($\mathcal{P}_T$; Ours). The supervised baselines miss populations outside their training distribution and collapse under a modality change (\textcolor{red}{dashed boxes}), whereas CoP recovers them from one click per type.}
  \label{fig-qualitative}
\end{figure}

The qualitative comparisons in \cref{fig-qualitative} and \cref{sec:qual} show these failure patterns directly, across both H\&E and non-H\&E benchmarks. The fully-supervised baselines~\cite{CellViT,CA-SAM2} leave whole cell populations unsegmented when the target lies outside their training distribution, and produce almost empty masks once the staining modality changes. The per-instance SAM3~\cite{SAM3} reference recovers these regions but requires one click for every cell ($\mathcal{P}_N$), whereas CoP reaches comparable coverage from one click per cell type ($\mathcal{P}_T$), and one click per image when types are unlabeled, by propagating each click through the frozen image encoder's features. Its masks therefore stay complete on both unseen cell types and unseen imaging modalities.

\subsection{Ablation Study}
\label{sec:ablation}

We ablate CoP's design choices on CoNIC~\cite{CoNIC} with SAM3~\cite{SAM3} as the frozen image encoder. Unless stated otherwise, every number below is AJI under this single setting, and the default CoP with all components enabled reaches 0.731, which is 91\% of the per-instance upper bound (0.801, \cref{tab-typed}). Removing any single component degrades this result substantially. We isolate each contribution below.

\begin{table*}[t]
\centering
\providecommand{\cmark}{\ding{51}}
\providecommand{\xmark}{\ding{55}}
\caption{\textbf{Ablation of CoP's two modules on CoNIC~\cite{CoNIC}} with SAM3~\cite{SAM3} as the frozen image encoder. (a)~HSG multi-scale gating ($S_h$, $S_l$). (b)~FPR selection rule over the clearance $d(c)=\min_{q\in\mathcal{Q}^{(t)}}\lVert c-q\rVert_2$ of a candidate to all prior prompts in \cref{eq:fpr}; $\tilde{d}$ is its median. $\Delta$ is AJI relative to the default CoP (\colorbox{gray!15}{shaded}).}
\label{tab:ablation}
\setlength{\tabcolsep}{5pt}
\begin{minipage}[b]{0.36\linewidth}
  \centering
  \fitbox{\begin{tabular}{cc c c}
    \toprule
    $S_h$ & $S_l$ & AJI\,$\uparrow$ & $\Delta$ \\
    \midrule
    \xmark & \cmark & 0.419 & \textcolor{red}{$-$0.312} \\
    \cmark & \xmark & 0.545 & \textcolor{red}{$-$0.186} \\
    \rowcolor{gray!15}
    \cmark & \cmark & \textbf{0.731} & --- \\
    \bottomrule
  \end{tabular}}
  \\[3pt]{\small (a)~HSG gating}
\end{minipage}
\hfill
\begin{minipage}[b]{0.62\linewidth}
  \centering
  \fitbox{\begin{tabular}{l c c c}
    \toprule
    Rule & Equation & AJI\,$\uparrow$ & $\Delta$ \\
    \midrule
    Closest  & $\argmin_{c} d(c)$ & 0.612 & \textcolor{red}{$-$0.119} \\
    Midpoint & $\argmin_{c}\lvert d(c)-\tilde{d}\rvert$ & 0.658 & \textcolor{red}{$-$0.073} \\
    \rowcolor{gray!15}
    Farthest & $\argmax_{c} d(c)$ & \textbf{0.731} & --- \\
    \bottomrule
  \end{tabular}}
  \\[3pt]{\small (b)~FPR selection}
\end{minipage}
\end{table*}

\begin{figure*}[t]
  \centering
  \includegraphics[width=0.70\linewidth]{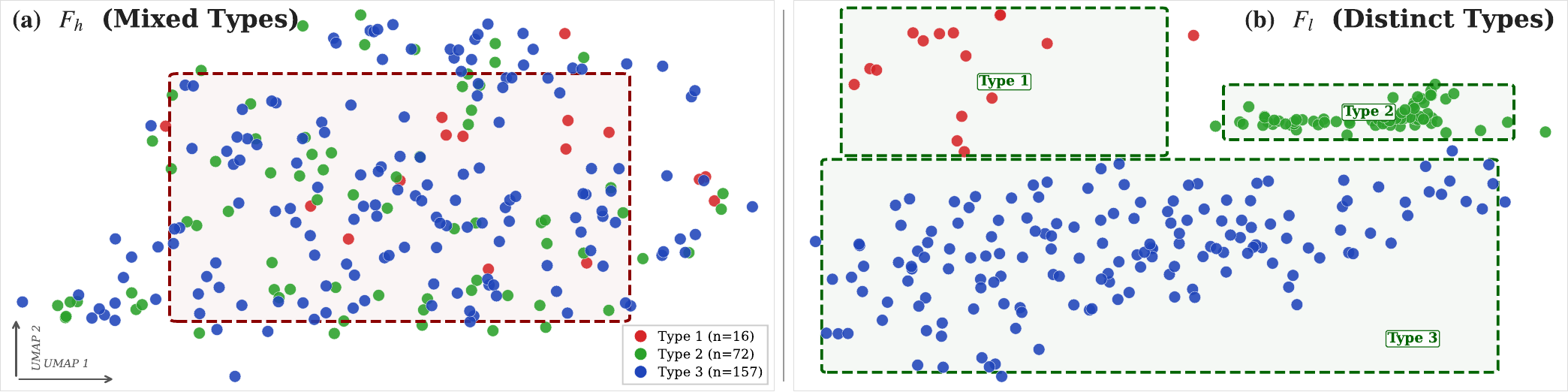}
  \vspace{-0.7em}
  \caption{
  {UMAP \cite{mcinnes2018umap} of SAM's frozen image encoder features at GT instance centroids. The UMAP embeddings are extracted from the input image used in \cref{fig_highlight}. (a)~$F_h$ mixes cell types; (b)~$F_l$ groups same-type cells without any training. Its dataset-level counterpart is the $S_l$ ablation of \cref{tab:ablation}a.}
  }
  \label{fig:tsne}
\end{figure*}

\textbf{Multi-scale similarity gating.} Gating $S_h$ with $S_l$ is essential (\cref{tab:ablation}a). Using $S_h$ alone lowers AJI to 0.545 (\textcolor{red}{$\downarrow$0.186}), as high-resolution features also respond to tissue of similar texture and these false positives compound through the recursion, so precision drops below 0.60 by $t{=}15$. Using $S_l$ alone lowers it to 0.419 (\textcolor{red}{$\downarrow$0.312}), since its coarse resolution localizes prompts poorly. The gated product $S_h\odot S_l$ instead holds precision above 0.96 at every iteration at comparable recall, because $F_l$, drawn from deeper layers with a larger receptive field, clusters cells by semantic identity (\cref{fig:tsne}), whereas $F_h$ pinpoints centers but with high semantic uncertainty. Gating the two therefore suppresses the spatial noise of $F_l$ and the semantic uncertainty of $F_h$.

\textbf{Adaptive thresholding.} HSG binarizes $S_h\odot S_l$ at $\tau=\mu+\sigma$ rather than at a constant. A fixed cutoff cannot follow the similarity statistics, which drift at every recursion as the prompt set grows: sweeping $\tau\in[0.1,0.9]$ spans AJI 0.247 to 0.585 (up to \textcolor{red}{$\downarrow$0.484}), since any constant value is at once too permissive on some iterations and too strict on others. Removing $\sigma$ and cutting at the mean $\mu$ alone admits roughly half of all candidates, so weakly reliable points between $\mu$ and $\mu+\sigma$ enter the reliable set and accumulate false positives across recursions, lowering AJI to 0.641 (\textcolor{red}{$\downarrow$0.090}). Retaining $\mu+\sigma$ adapts the cutoff to each iteration's distribution and preserves a high-recall yet high-precision group throughout; this data-adaptive rule follows the confidence gating of COIN~\cite{COIN}, which we adopt directly.

\textbf{FPR selection.} All variants share the clearance $d(c)$ of \cref{eq:fpr} and differ only in how they select the next prompt from it (\cref{tab:ablation}b). Our farthest rule, the $\argmax_{c} d(c)$ of \cref{eq:fpr}, reaches 0.731. Replacing that $\argmax$ with $\argmin_{c} d(c)$ (closest) or with $\argmin_{c}\lvert d(c)-\tilde{d}\rvert$ (midpoint, where $\tilde{d}$ is the median clearance) lowers AJI to 0.612 (\textcolor{red}{$\downarrow$0.119}) and 0.658 (\textcolor{red}{$\downarrow$0.073}), respectively. Both alternatives keep sampling near already-covered regions and spend iterations re-confirming known cells, whereas maximizing the clearance drives each new prompt into unexplored tissue and resolves the spatial-coverage bottleneck of HSG (\cref{sec:hsg}).

\textbf{Recursive propagation.} Removing FPR (\cref{sec:fpr}) restricts HSG (\cref{sec:hsg}) to the neighborhood of the initial click, and AJI drops from 0.731 to 0.482 (\textcolor{red}{$\downarrow$0.249}). Feature similarity to a single prompt decays across distant tissue, so one round of gating recovers only nearby cells and leaves the rest of the field uncovered. The iterative expansion defined in \cref{eq:fpr}, which re-prompts from newly discovered cells, is what propagates the initial click to distant instances.

\subsection{Additional Qualitative Results}
\label{sec:qual}
\begin{figure}[p]
\centering
\includegraphics[width=\linewidth]{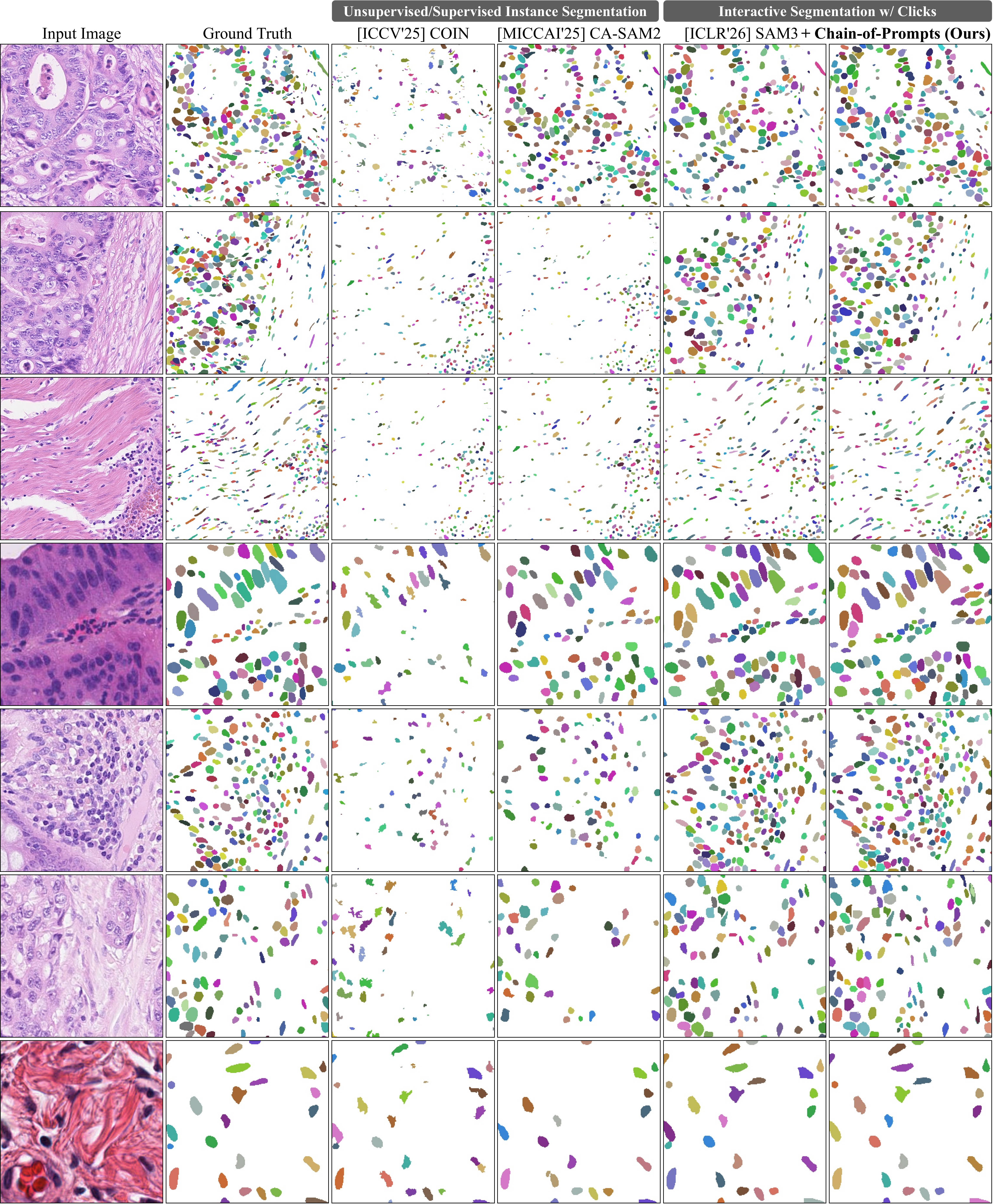}
\caption{\textbf{Qualitative comparison on H\&E benchmarks with cell-type annotations} (CoNIC~\cite{CoNIC}, CoNSeP~\cite{CoNSeP}, PanNuke~\cite{PanNuke}). Columns: input, ground truth, COIN~\cite{COIN}, CA-SAM2~\cite{CA-SAM2}, SAM3~\cite{SAM3} ($\mathcal{P}_N$), SAM3 + CoP ($\mathcal{P}_T$).}
\label{fig:qual-he-typed}
\end{figure}

\begin{figure}[p]
\centering
\includegraphics[width=\linewidth]{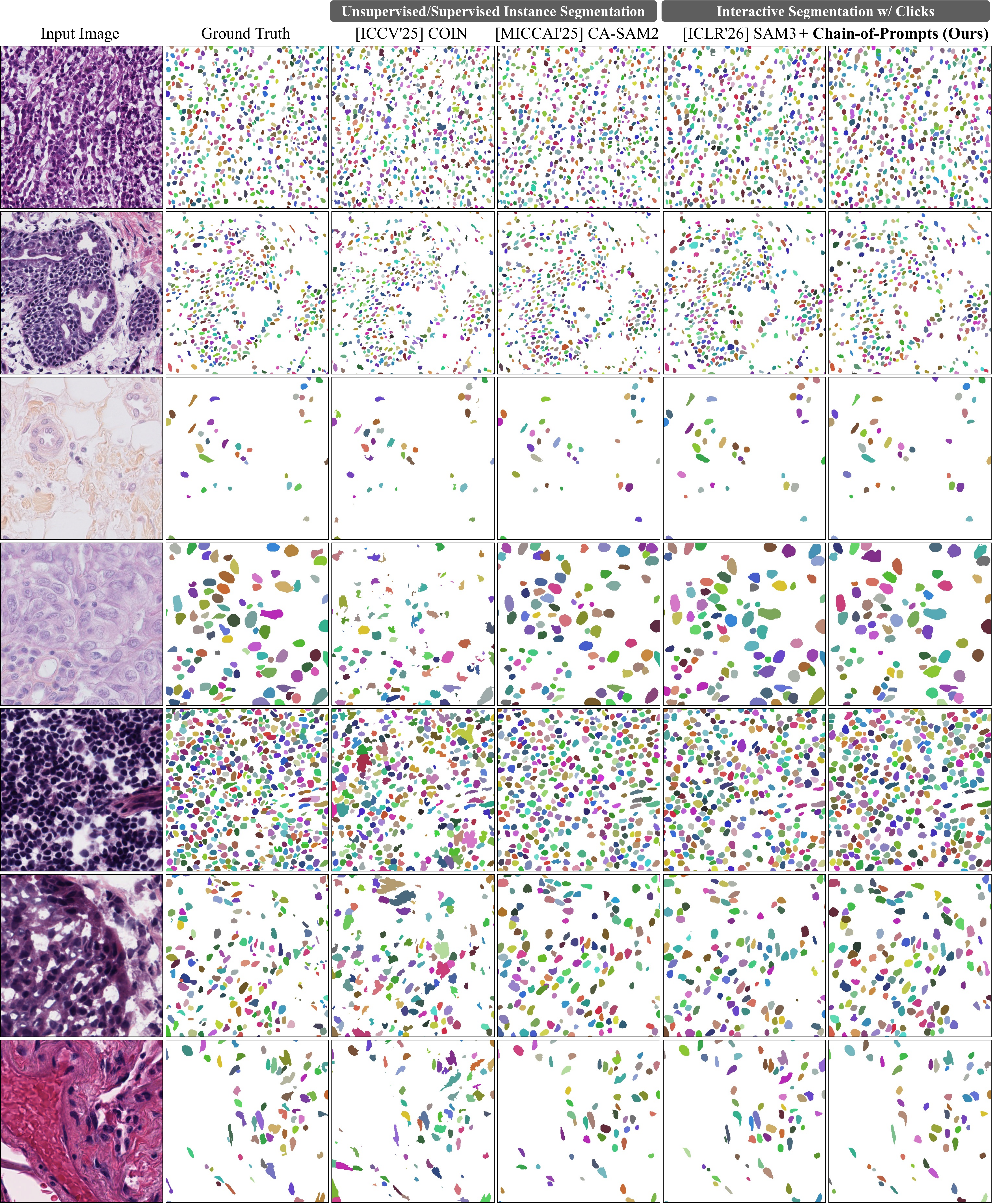}
\caption{\textbf{Qualitative comparison on H\&E benchmarks without cell-type annotations} (MoNuSeg~\cite{MoNuSeg}, TNBC~\cite{TNBC}, CryoNuSeg~\cite{CryoNuSeg}, CPM-17~\cite{CPM-17}). Columns: input, ground truth, COIN~\cite{COIN}, CA-SAM2~\cite{CA-SAM2}, SAM3~\cite{SAM3} ($\mathcal{P}_N$), SAM3 + CoP ($\mathcal{P}_T$).}
\label{fig:qual-he-untyped}
\end{figure}

\begin{figure}[p]
\centering
\includegraphics[width=\linewidth]{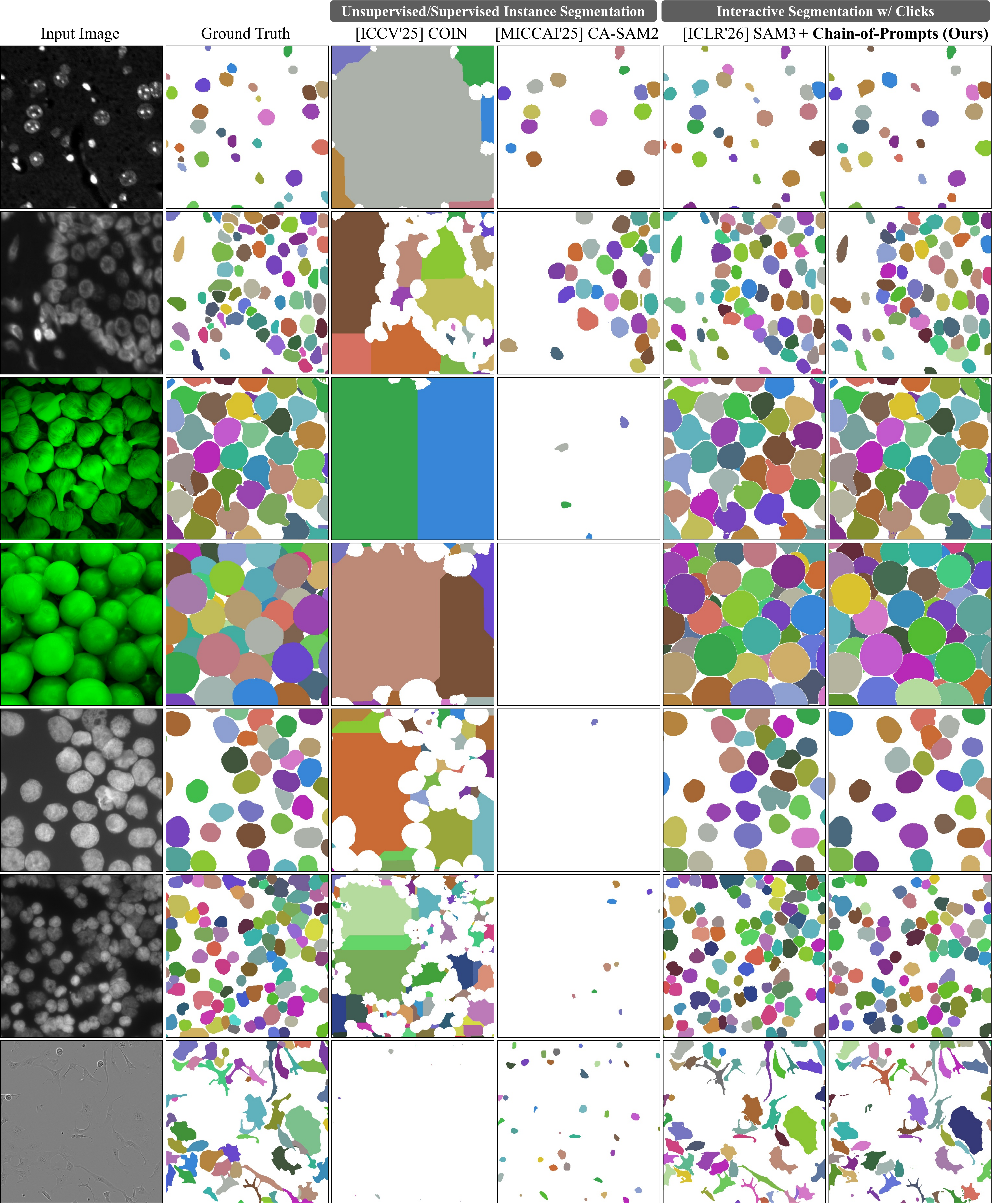}
\caption{\textbf{Qualitative comparison on non-H\&E benchmarks without cell-type annotations} (CellBinDB~\cite{CellBinDB}, Cellpose~\cite{Cellpose}, Kromp~\cite{Kromp}, LIVECell~\cite{LIVECell}). Columns: input, ground truth, COIN~\cite{COIN}, CA-SAM2~\cite{CA-SAM2}, SAM3~\cite{SAM3} ($\mathcal{P}_N$), SAM3 + CoP ($\mathcal{P}_T$).}
\label{fig:qual-nonhe}
\end{figure}

\Cref{fig-qualitative} compares CoP with the fully-supervised baselines and the per-instance reference across both H\&E and non-H\&E benchmarks, and \cref{fig:qual-he-typed,fig:qual-he-untyped,fig:qual-nonhe} extend the comparison to the three settings of \cref{tab:datasets}, each adding an unsupervised COIN~\cite{COIN} column beside the per-instance upper bound. With cell-type annotations (\cref{fig:qual-he-typed}), COIN and the fully-supervised CA-SAM2~\cite{CA-SAM2} recover only a fraction of the instances, whereas SAM3 + CoP recovers most of the per-instance SAM3~\cite{SAM3} coverage from one click per cell type. Without cell-type annotations (\cref{fig:qual-he-untyped}), a single click per image suffices to track the per-instance masks. Under a staining-modality shift (\cref{fig:qual-nonhe}), CA-SAM2 collapses to near-empty masks while CoP preserves coverage on frozen encoder features without any gradient update. Consistent with \cref{tab-typed,tab-wotype,tab-nonhe}, CoP stays close to the upper bound because it reads instance structure already present in the frozen image encoder rather than a learned appearance prior.

\section{Discussion}
\label{sec:discussion}

\subsection{Prompt Propagation}
\label{sec:propagation}

\begin{table}[!ht]
\centering
\setlength{\tabcolsep}{5pt}
\caption{\textbf{Effect of the propagation rule on the cell-type-annotated benchmarks~\cite{CoNIC,CoNSeP,PanNuke}.} PerSAM~\cite{PerSAM} and Matcher~\cite{Matcher} are one-shot \emph{semantic} methods, so we port each to instance segmentation on the same frozen SAM3~\cite{SAM3}, keeping every pseudo prompt as one instance, under the hyperparameters of its own paper (the ports are detailed in the text).}
\label{tab:seedrule}
\fitbox{\begin{tabular}{l c|cc|cc|cc}
\toprule
\multirow{2}{*}{Method} & \multirow{2}{*}{Prompt}
  & \multicolumn{2}{c|}{\textbf{CoNIC}}
  & \multicolumn{2}{c|}{\textbf{CoNSeP}}
  & \multicolumn{2}{c}{\textbf{PanNuke}} \\
 & & AJI & Dice & AJI & Dice & AJI & Dice \\
\midrule
\ub{SAM3 \cite{SAM3}} & \ub{$\mathcal{P}_N$} & \ub{0.801} & \ub{0.938} & \ub{0.716} & \ub{0.867} & \ub{0.830} & \ub{0.975} \\
SAM3 + PerSAM \cite{PerSAM} & $\mathcal{P}_T$ & 0.036 & 0.097 & 0.007 & 0.015 & 0.113 & 0.221 \\
SAM3 + Matcher \cite{Matcher} & $\mathcal{P}_T$ & 0.053 & 0.239 & 0.021 & 0.089 & 0.202 & 0.400 \\
\rowcolor{gray!15}
\textbf{SAM3 + CoP (Ours)} & $\mathcal{P}_T$ & \textbf{0.731} & \textbf{0.898} & \textbf{0.653} & \textbf{0.831} & \textbf{0.782} & \textbf{0.932} \\
\bottomrule
\end{tabular}}
\end{table}

Any similarity-based rule could in principle stand in for CoP's, so we swap that component alone and keep the frozen SAM3~\cite{SAM3} encoder, the initial click and the mask assembly fixed (\cref{tab:seedrule}). PerSAM~\cite{PerSAM} and Matcher~\cite{Matcher} were built for one-shot semantic transfer between two images, where the answer is a single region, and they carry that design over: each reads similarity from one feature map, fixes in advance how many locations it will return, and propagates once without revisiting. CoP differs on all three counts, gating two feature scales against each other, deriving its cutoff from the activation statistics of each step, and re-prompting until nothing new appears. The metrics separate these effects: AJI falls further than Dice because the masks that are produced merge touching cells, and Dice is low as well because propagation never leaves the neighborhood of the click, so most of the field is never prompted.

\textbf{Porting PerSAM and Matcher.} Neither method runs off the shelf here, because both target one-shot \emph{semantic} segmentation between a reference image and a target image and return one region rather than per-object masks. We therefore re-implement each as a prompt propagation rule inside a single image and keep every pseudo prompt it produces as one instance, sharing the rest of the pipeline with CoP so that only that rule differs: the reference is the same single click, the propagation features and the mask decoder are the frozen SAM3~\cite{SAM3}, and the click selection is computed once and reused, so every row starts from identical clicks. Hyperparameters follow each paper, namely PerSAM's~\cite{PerSAM} two-step cascaded post-refinement and Matcher's~\cite{Matcher} eight $k$-means clusters with a purity threshold of $0.02$. Three components could not be carried over. PerSAM's low-resolution mask feedback is backend-specific, so both refinement rounds re-prompt with the points and the bounding box of the current mask, which is the paper's second step, and its target-guided attention and target-semantic prompting act inside the decoder, where they refine a mask but cannot change how many masks an $\argmax$ prior yields. Matcher's patch-level bidirectional matching is vacuous within one image, since similarity is then symmetric, so the reverse test is applied to each cluster's mean feature instead, and its earth-mover ranking exists to select fragments for a single region, which the instance adaptation removes. Matching also reads the same SAM3 features rather than the separate DINOv2~\cite{DINOv2} encoder of the original, which holds the encoder fixed across every row.

\subsection{Propagation at a Fixed Click Budget}
\label{sec:budget}

\begin{table}[!ht]
\centering
\setlength{\tabcolsep}{6pt}
\caption{\textbf{Interaction cost against compute cost on CoNSeP~\cite{CoNSeP}.} AMG is automatic mask generation, the zero-click grid protocol of SAM~\cite{SAM1}; prompt symbols follow \cref{tab-typed}. Decodes/img counts mask-decoder calls, which for CoP is one per discovered cell. Latency is 0.6\,s of encoding plus 4.74\,ms per decoded prompt; ``CoP only'' is the HSG and FPR propagation on top of it (1.08\,s per click). All rows share the frozen SAM3~\cite{SAM3} encoder, decoder and assembly.}
\label{tab:budget}
\fitbox{\begin{tabular}{l c c c c c c c}
\toprule
\multirow{2}{*}{Method} & \multirow{2}{*}{Prompt} & \multirow{2}{*}{Clicks/img} & \multirow{2}{*}{Decodes/img} & \multicolumn{2}{c}{Latency/img} & \multirow{2}{*}{AJI\,$\uparrow$} & \multirow{2}{*}{Dice\,$\uparrow$} \\
 & & & & enc.+dec. & CoP only & & \\
\midrule
SAM3~\cite{SAM3} + AMG ($32{\times}32$) & \xmark & 0 & 1{,}024 & 5.45\,s & -- & 0.209 & 0.497 \\
SAM3~\cite{SAM3} + AMG ($64{\times}64$) & \xmark & 0 & 4{,}096 & 20.02\,s & -- & 0.250 & 0.559 \\
\midrule
\ub{SAM3~\cite{SAM3}} & \ub{$\mathcal{P}_N$} & \ub{626.9} & \ub{626.9} & \ub{3.57\,s} & \ub{--} & \ub{0.716} & \ub{0.867} \\
SAM3~\cite{SAM3} & $\mathcal{P}_T$ & 3.86 & 3.86 & 0.62\,s & -- & 0.007 & 0.013 \\
\rowcolor{gray!15}
\textbf{SAM3 + CoP (Ours)} & $\mathcal{P}_T$ & 3.86 & 566.7 & 3.29\,s & 4.17\,s & \textbf{0.653} & \textbf{0.831} \\
\bottomrule
\end{tabular}}
\end{table}

The $O(T)$ of group prompting is a claim about user interaction, not about computation. \Cref{tab:budget} separates the two on CoNSeP~\cite{CoNSeP}, the densest split in our evaluation. From 3.86 clicks CoP reaches 566.7 decoded prompts per image, 90.4\% of the 626.9 a per-instance user would place. This prompt-count ratio is distinct from the type-aware recall in \cref{tab:detection}. Encoding plus decoding therefore rises from 0.62\,s to 3.29\,s and the propagation itself adds 4.17\,s, while the manual budget falls by more than two orders of magnitude.

The same table separates what a CoP row and a per-instance row in \cref{tab-typed} differ in: how many clicks are spent, and whether they propagate. Spending CoP's own clicks without propagating reaches AJI 0.007, against 0.653 with it, so the budget explains none of the gap. At 3.86 clicks against 626.9 instances, one mask per click is the ceiling, and this row is the floor that recursion lifts. It also anchors \cref{tab:seedrule}: PerSAM~\cite{PerSAM} scores 0.007 here too, sitting at that floor by construction rather than through any weakness of our port, since a single $\argmax$ prior returns one instance per click.

Removing the click relocates the cost rather than removing it. A $32{\times}32$ grid decodes 1{,}024 points per image and a $64{\times}64$ grid decodes 4{,}096, where one click per cell type decodes 3.86: $265\times$ and $1{,}061\times$ that budget, and heavier than prompting all 626.9 annotated instances individually. The protocols that ask nothing of the user are therefore the most expensive to run, and they are also the only ones that cannot be aimed, since a grid has no way to express which cell population is wanted. CoP pays $147\times$ that budget on the same 3.86 clicks, still under the $162\times$ of the per-instance reference, because it decodes once per cell it has located rather than once per position a cell might occupy. Density does not close the gap either: quadrupling the grid lifts AJI only from 0.209 to 0.250, recovering 453 of the 626.9 nuclei instead of 292, so what a grid lacks is not resolution but any signal telling it where cells are.

\begin{table}[!ht]
\centering
\setlength{\tabcolsep}{6pt}
\caption{\textbf{One click per type against one click per image.} AJI on the three type-annotated
benchmarks with SAM3~\cite{SAM3} as the frozen encoder, reported as mean$\pm$std over at least ten
runs; for $\mathcal{P}_1$ the single click is sampled ten or more times from each type present in
the image. Retention is $\mathcal{P}_1$ as a percentage of $\mathcal{P}_T$, and $1/T$ uses the
per-image type counts of \cref{tab:datasets}.}
\label{tab:oneclick}
\fitbox{\begin{tabular}{l c c c c}
\toprule
Method & Prompt & CoNIC~\cite{CoNIC} & CoNSeP~\cite{CoNSeP} & PanNuke~\cite{PanNuke} \\
\midrule
\rowcolor{gray!15}
\textbf{SAM3 + CoP (Ours)} & $\mathcal{P}_T$ & \textbf{0.731}$\pm$0.005 & \textbf{0.653}$\pm$0.004 & \textbf{0.782}$\pm$0.002 \\
SAM3 + CoP & $\mathcal{P}_1$ & 0.209$\pm$0.006 & 0.175$\pm$0.006 & 0.437$\pm$0.004 \\
\midrule
Retention & & 28.6\% & 26.8\% & 55.9\% \\
$1/T$ & & 25.8\% & 25.9\% & 48.8\% \\
\bottomrule
\end{tabular}}
\end{table}

\Cref{tab:oneclick} asks whether the per-type budget is necessary at all, since a single click per image already suffices on the untyped benchmarks (\cref{tab-wotype,tab-nonhe}). It is: spending one click per image on type-annotated data costs 0.522, 0.478 and 0.345 AJI on CoNIC~\cite{CoNIC}, CoNSeP~\cite{CoNSeP} and PanNuke~\cite{PanNuke}. What remains is close to the $1/T$ share of the image, so one click recovers one cell population rather than the whole field, which is exactly the behavior the $O(T)$ reading predicts. The untyped benchmarks escape this because their instances are morphologically homogeneous and act as a single effective type, so one click already covers the only population present.

\subsection{Frozen Image Encoder}
\label{sec:encoder}

\begin{table}[!ht]
\centering
\setlength{\tabcolsep}{6pt}
\caption{\textbf{Frozen image encoder of CoP on CoNIC~\cite{CoNIC}.} HSG (\cref{sec:hsg}) and FPR (\cref{sec:fpr}) are held fixed and only the encoder used for propagation is swapped, while masks are still decoded by SAM3~\cite{SAM3}. For a fair comparison, every non-interactive encoder uses a single feature map following its standard practice, with SDXL~\cite{SDXL} using the segmentation-oriented feature extraction of DiffCut~\cite{DiffCut}. \#Params is the image-encoder size and $\Delta$ is AJI relative to SAM3~\cite{SAM3}.}
\label{tab:encoder}
\fitbox{\begin{tabular}{l l c c c}
\toprule
Frozen encoder of CoP & Pretraining objective & \#Params & AJI\,$\uparrow$ & $\Delta$ \\
\midrule
EVA-02-CLIP-E~\cite{EVA-CLIP} & Image-Text Alignment & 5B & 0.328 & \textcolor{red}{$-$0.403} \\
DINOv3~\cite{DINOv3}          & Self-supervised Learning & 7B & 0.335 & \textcolor{red}{$-$0.396} \\
SDXL~\cite{SDXL}      & Text-to-Image Generation & 3B & 0.435 & \textcolor{red}{$-$0.296} \\
\rowcolor{gray!15}
SAM3~\cite{SAM3}              & Interactive Segmentation & 0.5B & \textbf{0.731} & --- \\
\bottomrule
\end{tabular}}
\end{table}

Holding HSG (\cref{sec:hsg}) and FPR (\cref{sec:fpr}) fixed and swapping only the frozen encoder used for propagation (\cref{tab:encoder}), scale is not the deciding factor: at 0.5B parameters SAM3~\cite{SAM3} is the smallest of the four encoders yet the only one that lifts CoP to 0.731, while DINOv3~\cite{DINOv3} at 7B, EVA-02-CLIP-E~\cite{EVA-CLIP} at 5B, and SDXL~\cite{SDXL} at 3B stay between 0.328 and 0.435. Despite their far larger scale, encoders pretrained without prompt-mask supervision cluster features only by low-level appearance, so a single click over-propagates across visually similar cells and tissue, whereas the prompt-mask-supervised SAM3~\cite{SAM3} clusters at the instance level, which is the property HSG and FPR consume. Because the mask decoder is held fixed at SAM3 in every row, the difference reflects the propagation features alone rather than any change in decoding quality. This identifies prompt-mask pretraining, not encoder capacity, as the source of the instance-level structure CoP exploits.

\textbf{Two-scale inputs for the alternative encoders.} The single feature map is also the best setting for each alternative rather than a handicap: giving them the same two-scale treatment as SAM3, at a matched feature-depth ratio, lowers AJI by a further 18\% for DINOv3~\cite{DINOv3} and 15\% for EVA-02-CLIP-E~\cite{EVA-CLIP}, because their training objectives sit at the last layer and leave the intermediate layers unsupervised, so combining scales injects noise instead of the spatial detail that SAM's decoder-facing features carry.

\textbf{Relation to CryoSAM.} The choice of encoder also separates CoP from CryoSAM~\cite{CryoSAM}, which propagates one prompt to every particle of a category in a tomogram without training, but reads its match from a separate self-supervised DINOv2~\cite{DINOv2} and decodes with SAM~\cite{SAM1}; the closest encoder we run, DINOv3~\cite{DINOv3}, reaches 0.335 against 0.731 above. Its match is also a single coarse-to-fine pass at a constant similarity cutoff with a preset prompt budget, and its recursion completes one particle across the planes of a volume rather than reaching further instances. \Cref{sec:ablation} varies the other two on this data: replacing $\mu{+}\sigma$ with a constant cutoff costs between 0.146 and 0.484 AJI depending on where that constant is placed, and dropping the recursion costs 0.249.

\subsection{Robustness to Click Location}
\label{sec:robustness}

\begin{table}[!ht]
\centering
\setlength{\tabcolsep}{6pt}
\caption{\textbf{Spread of every CoP entry over the initial click.} Each entry in \cref{tab-typed,tab-wotype,tab-nonhe} is a mean over at least ten independent runs, every run started from a click sampled uniformly inside a target cell; this table collects the AJI standard deviations behind those means. The eight untyped benchmarks are summarized as a group average, since one click per image is the entire protocol there.}
\label{tab:stds}
\fitbox{\begin{tabular}{l c l c}
\toprule
Group & Prompt & Method & AJI std \\
\midrule
\multirow{4}{*}{H\&E type-annotated} & \multirow{4}{*}{$\mathcal{P}_T$}
   & SAM3 + CoP, CoNIC~\cite{CoNIC}      & $\pm$0.005 \\
 & & SAM3 + CoP, CoNSeP~\cite{CoNSeP}    & $\pm$0.004 \\
 & & SAM3 + CoP, PanNuke~\cite{PanNuke}  & $\pm$0.002 \\
 & & $\mu$SAM + CoP, all three           & $\pm$0.004 \\
\midrule
H\&E untyped and non-H\&E & $\mathcal{P}_1$ & SAM3 + CoP and $\mu$SAM + CoP & $\pm$0.003 \\
\bottomrule
\end{tabular}}
\end{table}

A user clicks wherever is convenient inside a cell, not at the centroid an evaluation script would pick, so every CoP entry in \cref{tab-typed,tab-wotype,tab-nonhe} is reported as a mean over at least ten runs from independently sampled clicks rather than from one favorable initialization. The spread behind those means is uniformly small (\cref{tab:stds}): $0.731{\pm}0.005$, $0.653{\pm}0.004$, and $0.782{\pm}0.002$ AJI on the three typed benchmarks and $\pm0.003$ on average over the eight untyped ones, while $\mu$SAM + CoP stays within $\pm0.004$ throughout. The largest standard deviation among the per-type SAM3 results is 0.005 AJI, on CoNIC~\cite{CoNIC}; the smallest, 0.002, is on PanNuke~\cite{PanNuke}, whose images carry the fewest types and therefore offer the least room for a click to land on an ambiguous cell. Even the largest is far below the margins that separate CoP from the supervised and open-vocabulary baselines of \cref{tab-typed}, so none of the comparisons in this paper turns on where the initial click happened to fall: the result is essentially independent of where in a target cell the prompt lands.

\subsection{Cross-Type Leakage}
\label{sec:leakage}
Because AJI and Dice are cell-type-agnostic (\cref{sec:metrics}), the reported scores evaluate instance segmentation rather than cell-type classification, and ``one click per type'' sets a prompt budget, not a purity guarantee. When two types look alike, a single click occasionally recovers instances from a neighboring type as well. Since the prompt budget is nonetheless one click per cell type, we add two complementary measurements of this leakage. \Cref{tab:pertype} rescores the same masks under a cell-type constraint across all four frozen encoders on CoNIC~\cite{CoNIC}, the cell-type-annotated benchmark with the most cell types, and \cref{tab:detection} then scores cell-type-aware detection with SAM3~\cite{SAM3} on all three type-annotated benchmarks.

\begin{table}[!ht]
\centering
\setlength{\tabcolsep}{5pt}
\caption{\textbf{Cell-type-specific scoring on CoNIC~\cite{CoNIC}.} The right half adds one
constraint: a prediction, whose cell type is that of the click that produced it, may match a ground
truth instance only when their cell types agree. \ub{Gray} rows are per-instance upper bounds
($\mathcal{P}_N$), whose cell types are correct by construction, and each
\colorbox{gray!15}{shaded} row is CoP applied to the model above it.}
\label{tab:pertype}
{
\fitbox{\begin{tabular}{l c|cc|cc}
\toprule
\multirow{2}{*}{Frozen encoder} & \multirow{2}{*}{Prompt} & \multicolumn{2}{c|}{\textbf{Cell-type-agnostic}} & \multicolumn{2}{c}{\textbf{Cell-type-specific}} \\
 & & AJI\,$\uparrow$ & Dice\,$\uparrow$ & AJI\,$\uparrow$ & Dice\,$\uparrow$ \\
\midrule
\ub{SAM1-H~\cite{SAM1}} & \ub{$\mathcal{P}_N$} & \ub{0.755} & \ub{0.915} & \ub{0.755} & \ub{0.914} \\
\rowcolor{gray!15}
SAM1-H + CoP & $\mathcal{P}_T$ & 0.692 & 0.838 & 0.689 & 0.835 \\
\ub{SAM2.1-L~\cite{SAM2}} & \ub{$\mathcal{P}_N$} & \ub{0.769} & \ub{0.922} & \ub{0.768} & \ub{0.921} \\
\rowcolor{gray!15}
SAM2.1-L + CoP & $\mathcal{P}_T$ & 0.711 & 0.846 & 0.703 & 0.843 \\
\midrule
\ub{$\mu$SAM~\cite{uSAM}} & \ub{$\mathcal{P}_N$} & \ub{0.759} & \ub{0.912} & \ub{0.758} & \ub{0.911} \\
\rowcolor{gray!15}
$\mu$SAM + CoP & $\mathcal{P}_T$ & 0.652 & 0.791 & 0.648 & 0.785 \\
\ub{SAM3~\cite{SAM3}} & \ub{$\mathcal{P}_N$} & \ub{0.801} & \ub{0.938} & \ub{0.801} & \ub{0.938} \\
\rowcolor{gray!15}
\textbf{SAM3 + CoP (Ours)} & $\mathcal{P}_T$ & \textbf{0.731} & \textbf{0.898} & \textbf{0.729} & \textbf{0.891} \\
\bottomrule
\end{tabular}}}
\end{table}

\textbf{Cell-type-constrained rescoring.} The first measurement asks what the same masks are worth when the recovered cell type has to be right as well. We read the cell type of each ground truth instance from the CoNIC cell-type labels, give each prediction the cell type of the click that produced it, which for $\mathcal{P}_N$ is the annotated nucleus the click was placed on and for $\mathcal{P}_T$ the queried cell type, and forbid a match between two instances whose cell types differ. Instances are decoded as in \cref{sec:metrics} and unmatched ones still enter the union with their full area, so the two halves of \cref{tab:pertype} differ through forbidden matches alone.

The $\mathcal{P}_N$ rows are the control: their clicks sit on annotated nuclei, so their cell types are right by construction, and the two halves agree to within 0.001 AJI and 0.001 Dice. Against that reference CoP loses between 0.002 and 0.008 AJI and between 0.003 and 0.007 Dice ($0.731{\to}0.729$ AJI for SAM3 + CoP and $0.652{\to}0.648$ for $\mu$SAM + CoP), retaining 98.9\% to 99.7\% of its cell-type-agnostic score with the ordering of the encoders unchanged, so a click almost always recovers instances of the type it was placed on. What leakage there is stays confined to cell types that are morphologically similar; the remaining cell types differ enough in shape and stain that propagation through the frozen features does not cross into them, which is why one click per cell type scores almost the same whether the cell type is read or ignored.

\begin{table}[!ht]
\centering
\setlength{\tabcolsep}{5pt}
\caption{\textbf{Cell-type-aware detection on the three type-annotated benchmarks.} A prediction
counts as a match only when it overlaps a ground truth instance of the queried cell type by
IoU${>}0.5$; Precision, Recall and $F_1$ are averaged over the cell types present in each image.
\ub{Gray} columns are the per-instance upper bound ($\mathcal{P}_N$). $F_1$ kept is
$F_1(\mathcal{P}_T)/F_1(\mathcal{P}_N)$ and Prompts used is $T/N$ from \cref{tab:datasets}.}
\label{tab:detection}
\fitbox{\begin{tabular}{l ccc ccc cc}
\toprule
\multirow{2}{*}{Dataset} & \multicolumn{3}{c}{\ub{SAM3~\cite{SAM3} ($\mathcal{P}_N$)}} & \multicolumn{3}{c}{\textbf{SAM3 + CoP} ($\mathcal{P}_T$)} & $F_1$ & Prompts \\
 & \ub{Prec.} & \ub{Rec.} & \ub{$F_1$} & Prec. & Rec. & $F_1$ & kept\,$\uparrow$ & used\,$\downarrow$ \\
\midrule
CoNIC~\cite{CoNIC}     & \ub{0.961} & \ub{0.956} & \ub{0.958} & 0.972 & 0.868 & \textbf{0.917} & 95.7\% & 3.4\% \\
CoNSeP~\cite{CoNSeP}   & \ub{0.968} & \ub{0.918} & \ub{0.941} & 0.985 & 0.821 & \textbf{0.896} & 95.2\% & 0.6\% \\
PanNuke~\cite{PanNuke} & \ub{0.949} & \ub{0.942} & \ub{0.945} & 0.953 & 0.908 & \textbf{0.929} & 98.3\% & 8.4\% \\
\bottomrule
\end{tabular}}
\end{table}

\textbf{Cell-type-aware detection.} That reading has one blind spot: because rescoring only forbids matches, a cell CoP never reached and a cell it reached with the wrong cell type both count the same way. \Cref{tab:detection} separates the two by scoring detection rather than overlap, and repeats the measurement on CoNSeP~\cite{CoNSeP} and PanNuke~\cite{PanNuke} as well. Precision matches or exceeds the per-instance ceiling on every one of them (0.972, 0.985, and 0.953 against 0.961, 0.968, and 0.949 on CoNIC~\cite{CoNIC}, CoNSeP~\cite{CoNSeP}, and PanNuke~\cite{PanNuke}), because HSG (\cref{sec:hsg}) only decodes points that survive its gating, so a click almost never produces a mask of the wrong cell type. The one real gap is recall, and it narrows where more prompts are spent: PanNuke~\cite{PanNuke} uses $8.4\%$ of the per-instance prompts and loses $0.034$ recall, while CoNSeP~\cite{CoNSeP} uses $0.6\%$ and loses $0.097$. Even at these budgets ($3.4\%$, $0.6\%$, and $8.4\%$ of the per-instance clicks), type-aware $F_1$ stays above $95\%$ of the ceiling on all three benchmarks ($95.7\%$, $95.2\%$, and $98.3\%$), which is the same conclusion the cell-type-agnostic scores of \cref{tab-typed} reach through a different metric.

The leakage is harmless here, since CoP targets cell-type-agnostic segmentation and such an instance still yields a correct mask. Making CoP cell-type-specific would instead require suppressing this over-propagation with type-discriminative cues read from the same frozen image features, a direction we leave to future work.

\subsection{Limitations}
\label{sec:limitations}
CoP is bounded by its base model. Because it never trains on cell annotations and reuses the frozen encoder as is, it cannot recover cells that the per-instance upper bound itself misses: any instance SAM3~\cite{SAM3} fails to segment even from a correct point prompt (\cref{tab-typed,tab-wotype,tab-nonhe}) is also missed by CoP, an inherent consequence of our training-free design.

\section{Conclusion}
In this paper, we present \textbf{Chain-of-Prompts (CoP)}, a training-free framework that discovers and segments all same-type cells from a single user click by recursively propagating prompts through frozen SAM features. Our key finding is that SAM's frozen image encoder already clusters same-type cells in its multi-scale feature space before any prompt is given, and CoP exploits this property through non-parametric gating without additional training. Across eleven benchmarks spanning both H\&E and non-H\&E modalities, CoP retains over 90\% of SAM3's per-instance performance while requiring 97.6\% fewer clicks overall, generalizes to unseen cell types and imaging modalities without adaptation, and surpasses fully-supervised methods. By reducing hundreds of manual annotations to a single click per cell type, CoP shows that interactive foundation models can be leveraged far more efficiently than the current paradigm assumes, establishing group prompting as a practical way to cut annotation effort, which scales with cell types, not cell counts. Validating it with pathologists is left to future work.

\section*{Acknowledgements}
This work was partly supported by the KHIDI grant funded by the Korean government (MOHW) [No.RS-2025-02307233, No.RS-2026-25613012], the NRF or IITP grants funded by the Korean government (MSIT) [No.05-26-04-0094, No.RS-2026-25472075, No.RS-2025-02305581, No.RS-2025-25442338, and No.RS-2021-II211343], the ITIP grant funded by the Korean government (MOTIR) [No.RS-2026-25549946], the KOCCA grant funded by the Korean government (MCST) [No.RS-2026-25527029], the Research grant from SNU, and the Strategic Hub grant for International Research Collaboration of SNU. Kyungsu Kim is affiliated with the School of Transdisciplinary Innovations, Department of Biomedical Science, Interdisciplinary Program in AI (IPAI) \& Graduate School of AI, Medical Research Center, and AI Institute at SNU.

\bibliographystyle{plainnat}
\begingroup\clubpenalty=0 \widowpenalty=0 \displaywidowpenalty=0
\bibliography{main}
\endgroup

\end{document}